\documentclass{article}
\usepackage{hyperref}
\usepackage{url}
\usepackage{lastpage}

\usepackage{microtype}
\usepackage{graphicx}
\usepackage{subfigure}
\usepackage{booktabs} % for professional tables
\usepackage{amsmath}
\usepackage{amssymb}
\usepackage{amsthm}
\usepackage{xcolor}
\usepackage{overpic}
\usepackage[toc,page]{appendix}
\usepackage[margin=1in]{geometry}

\usepackage[numbers,sort&compress]{natbib}
\usepackage{bibunits}
\defaultbibliographystyle{natbib}
\usepackage[compact]{titlesec}

\newcommand{\be}{\begin{eqnarray}}
\newcommand{\ee}{\end{eqnarray}}

\newcommand{\cC}{\ensuremath{\mathcal{C}}}

\newcommand{\cK}{\ensuremath{\mathcal{K}}}
\newcommand{\cM}{\ensuremath{\mathcal{M}}}
\newcommand{\cN}{\ensuremath{\mathcal{N}}}
\newcommand{\cO}{\ensuremath{\mathcal{O}}}
\newcommand{\cP}{\ensuremath{\mathcal{P}}}

\newcommand{\es}[2] {\begin{equation} \label{#1} \begin{split} #2 \end{split} \end{equation}}

\DeclareMathOperator{\trace}{Tr}

\newcommand{\lexpp}[1]{\mathbb{E}_{#1}\left[}
\newcommand{\rexp}{\right]}

\newcommand{\op}{overparameterized}
\newcommand{\up}{underparameterized}

\newtheorem{thm}{Theorem}
\newtheorem{supp_thm}{Theorem}
\newtheorem{lemma}{Lemma}

\usepackage{authblk}
\font\authfont=cmr12 at 10pt
\date{\href{mailto:yasamanb@google.com}{\small yasamanbahri@gmail.com}, \href{mailto:edyer@google.com}{\small edyer@google.com}, \href{mailto:jaredk@jhu.edu}{\small jaredk@jhu.edu}, \href{mailto:jaehlee@google.com}{\small jaehlee@google.com}, \href{mailto:usharma7@jhu.edu}{\small usharma7@jhu.edu}}
\begin{document}

%\twocolumn[

\title{\textbf{Explaining Neural Scaling Laws}}

\author[1]{\authfont Yasaman Bahri\thanks{All authors contributed to all aspects of this work.}}
\author[1]{\authfont  Ethan Dyer*}
\author[2]{\authfont  Jared Kaplan*}
\author[1]{\authfont  Jaehoon Lee*}
\author[2]{\authfont  Utkarsh Sharma*\thanks{A portion of work completed during an internship at Google.}}
\affil[1]{Google DeepMind, Mountain View, CA}
\affil[2]{Department of Physics and Astronomy,
Johns Hopkins University}

\maketitle
\begin{abstract}
\noindent
The population loss of trained deep neural networks often follows precise power-law scaling relations with either the size of the training dataset or the number of parameters in the network. We propose a theory that explains the origins of and connects these  scaling laws. We identify \emph{variance-limited} and \emph{resolution-limited} scaling behavior for both dataset and model size, for a total of four scaling regimes. The variance-limited scaling follows simply from the existence of a well-behaved infinite data or infinite width limit, while the resolution-limited regime can be explained by positing that models are effectively resolving a smooth data manifold. In the large width limit, this can be equivalently obtained from the spectrum of certain kernels, and we present evidence that large width and large dataset resolution-limited scaling exponents are related by a duality. We exhibit all four scaling regimes in the controlled setting of large random feature and pretrained models and test the predictions empirically on a range of standard architectures and datasets. We also observe several empirical relationships between datasets and scaling exponents under modifications of task and architecture aspect ratio. Our work provides a taxonomy for classifying different scaling regimes, underscores that there can be different mechanisms driving improvements in loss, and lends insight into the microscopic origins of and relationships between scaling exponents. 
\end{abstract}

\begin{figure*}[h]
     \hspace{0.5cm}
     \begin{subfigure}
        \centering
        \includegraphics[width=1.0\textwidth]{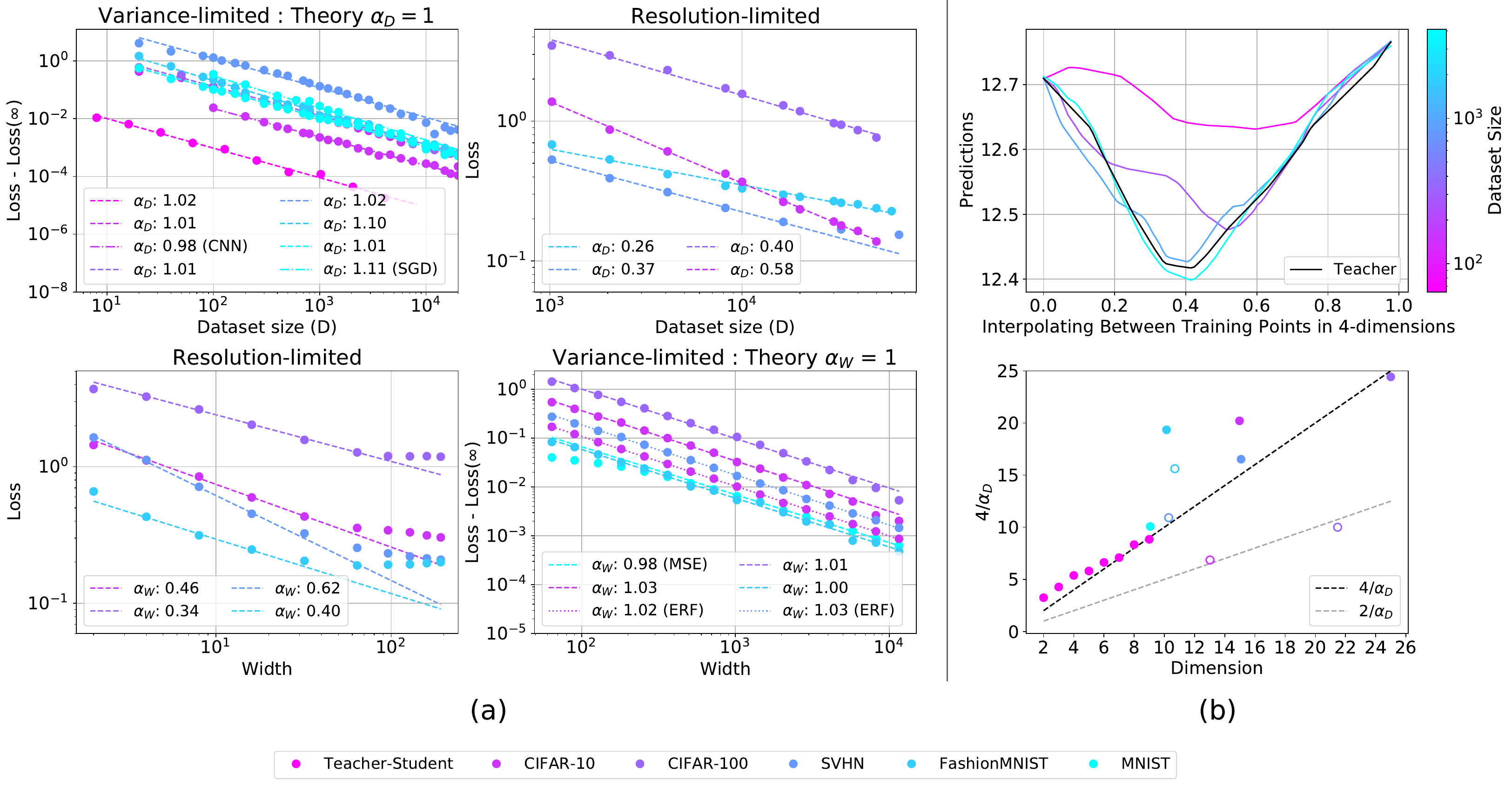}
         %\caption{}
         \label{fig:UP_dataset}
     \end{subfigure}
     \hfill
     \vspace{-0.5cm}
     \caption{\small\textbf{(a) Four scaling regimes.} Here we exhibit the four regimes we focus on in this work. \textbf{(top-left, bottom-right)} \emph{Variance-limited} scaling of \up{} models with dataset size and \op{} models with number of parameters (width) exhibit universal scaling ($\alpha_{D}=\alpha_{W}=1$) independent of the architecture or underlying dataset. \textbf{(top-right, bottom-left)} \emph{Resolution-limited} \op{} models with dataset or \up{} models with model size exhibit scaling  with exponents that depend on the details of the data distribution.
     These four regimes are also found in random feature (Figure~\ref{fig:kernel_banner}a) and pretrained models (see supplement).
     \textbf{(b) Resolution-limited models interpolate the data manifold.} Linear interpolation between two training points in a four-dimensional input space \textbf{(top)}.  We show a teacher model and four student models, each trained on different sized datasets. In all cases teacher and student approximately agree on the training endpoints, but as the training set size increases they increasingly match everywhere. \textbf{(bottom)} We show $4/\alpha_{D}$ versus the data manifold dimension (input dimension for teacher-student models, intrinsic dimension for standard datasets). We find that the teacher-student models follow the $4/\alpha_{D}$ (dark dashed line), while the relationship for a four layer convolutional neural network (solid) and Wide ResNet architecture (hollow) on standard datasets is less clear.
    }
        \label{fig:four_regimes}
        \label{fig:visualizing_interpolation}
        \label{fig:data_manifold_intro}
\end{figure*}

For a large variety of models and datasets, neural network performance has been empirically observed to scale as a power law with model size and dataset size~\citep{hestness2017deep, kaplan2020scaling, rosenfeld2020a, henighan2020scaling}. These exponents determine how quickly performance, as measured by the population loss, improves with more data and larger models. We would like to understand why these power laws emerge. For example, what features of the data and models determine the values of the power-law exponents? Is there a taxonomy behind scaling regimes, where regimes are governed by different underlying mechanisms? Are dataset and model size scaling exponents related in any way? And finally, which aspects of scaling behavior might exhibit universal signatures, and which aspects are strongly dependent on the ``microscopic" aspects of the problem? A theoretically and empirically-grounded understanding of these questions could provide guidance for machine learning in the modern era of large models and training data.

In this work, we present a theoretical framework for understanding scaling laws in trained deep neural networks. 
We identify four related scaling regimes with respect to the number of model parameters $P$ and the dataset size $D$.  With respect to each of $D$, $P$, we define both a \emph{variance-limited} regime and a \emph{resolution-limited} regime.

\section{Scaling laws for neural networks}

\subsection{Variance-limited regime}
In the limit of infinite data or an arbitrarily wide model, some aspects of neural network training simplify.  Specifically, if we fix one of $D,P$ and study scaling with respect to the other parameter as it becomes arbitrarily large, then the difference between the finite test loss and its limiting value scales as $1/x$, i.e. as a power law with exponent $1$, with $x = D$ or $\sqrt{P} \propto$ width in deep networks and $x = D$ or $P$ in linear models.  

\subsection{Resolution-limited regime}

In this regime,  one of $D$ or $P$ is effectively infinite, and  we study scaling as the \emph{other} parameter increases.  In this case, a variety of works have empirically observed power-law scalings  $1/x^\alpha$, typically with $0 < \alpha < 1$ for both $x = P$ or $D$. We derive exponents in this regime precisely in the setting of random feature models (c.f. next section). Empirically, we find that our theoretical predictions for exponents hold in pretrained, fine-tuned models even though these lie outside our theoretical setting.

For more general nonlinear models, we propose a refinement of naive bounds into estimates via expansions that hold asymptotically. These rely on the idea that additional data (in the infinite model-size limit) or added model parameters (in the infinite data limit) are used by the model to carve up the data manifold into smaller components. For smooth manifolds, loss, and network, the test loss will depend on the linear size of a sub-region, while it is the $d$-dimensional sub-region volume that scales inversely with $P$ or $D$, giving rise to $\alpha \propto 1/d$.\footnote{A visualization of this successively better approximation with dataset size is shown in Figure~\ref{fig:data_manifold_intro}b for models trained to predict data generated by a random fully-connected network.} To test this empirically, we make measurements of the resolution-limited exponents in neural networks and intrinsic dimension of the data manifold, shown in Figure~\ref{fig:data_manifold_intro}b.

\subsection{Explicit derivation} We derive the scaling laws for these four regimes explicitly in the setting of random feature teacher-student models, which also applies to neural networks in the large width limit. This setting allows us to solve for the test error directly in terms of the feature covariance (kernel). The scaling of the test loss then follows from the asymptotic decay of the spectrum of the covariance matrix. For generic continuous kernels on a $d$-dimensional manifold, we can further relate this to the dimension of the data manifold.

\subsection{Summary of contributions}

\begin{enumerate}
    \item We propose four scaling regimes for neural networks. The variance-limited and resolution-limited regimes originate from different mechanisms, which we identify. To our knowledge, this categorization has not been previously exhibited. We provide empirical support for all four regimes in deep networks on standard datasets. 
    \item We derive the variance-limited regime under simple yet general assumptions (Theorem \ref{thm:loss_var}). 
    \item We present a hypothesis for resolution-limited scaling through refinement of naive bounds (Theorems \ref{thm:interp_data}, \ref{thm:interp_params}), for general nonlinear models. We empirically test the dependence of the estimates on intrinsic dimension of the data manifold for deep networks on standard datasets (Figure~\ref{fig:data_manifold_intro}b).
    \item In the setting of random feature teacher-student networks, we derive \emph{both} variance-limited and resolution-limited scaling exponents exactly. In the latter case, we relate this to the spectral decay of kernels. We identify a novel \emph{duality} that exists between model and dataset size scaling. 
    \item We empirically investigate predictions from the random features setting in pretrained, fine-tuned models on standard datasets and find they give excellent agreement.
    \item We study the dependence of the scaling exponent on changes in architecture and data, finding that (i) changing the input distribution via switching datasets and (ii) the addition of noise have strong effects on the exponent, while (iii) changing the target task via superclassing does not.
\end{enumerate}

\subsection{Related works}
There have been a number of recent works demonstrating empirical scaling laws \citep{hestness2017deep, kaplan2020scaling, rosenfeld2020a, henighan2020scaling, rosenfeld2021predictability} in deep neural networks, including scaling laws with model size, dataset size, compute, and other observables such as mutual information and pruning.  Some precursors \citep{ahmad1989scaling, cohn1991} can be found in earlier literature. Recently, scaling laws have also played a significant role in motivating work on the largest models that have yet been developed \citep{brown2020language, hoffmann2022empirical}.  

There has been comparatively little work on theoretical ideas \citep{sharma2020neural, bisla2021theoretical} that match and explain empirical findings in generic deep neural networks.
 In the particular case of large width, deep neural networks behave as random feature models \citep{nealthesis, lee2018deep, matthews2018, Jacot2018ntk, lee2019wide, Dyer2020Asymptotics}, and known results on the loss scaling of kernel methods can be applied \citep{spigler2020asymptotic, bordelon2020spectrum}. Though not in the original, \cite{bordelon2020spectrum} analyze resolution-limited dataset size scaling for power-law spectra in later versions. The decay of test error in ridge regression under certain settings has been studied in prior work, including \citep{caponnetto2007, steinwart2009, fischer2020}.
 
 During the completion of this work, \cite{hutter2021learning} presented a specific solvable model of learning exhibiting non-trivial power-law scaling for power-law (Zipf) distributed features. This does not directly relate to the settings studied in this work, or present bounds that supersede our results.  
 Concurrent to our work, \cite{bisla2021theoretical} presented a derivation of the resolution-limited scaling with dataset size, also stemming from nearest-neighbor distance scaling on data manifolds. However, they do not discuss requirements on model versus dataset size or how this scaling behavior fits into other asymptotic scaling regimes. A few recent works, appearing after the completion of this manuscript, also investigate the scaling of test error in related settings. \cite{cui2021generalization} studies the decay of test error with dataset size for kernel regression in a  high-dimensional limit with Gaussian design. \cite{maloney2022solvable} examine further a teacher-student framework similar to ours, deriving joint scaling laws using techniques from random matrix theory.  Finally, \cite{wei2022more} theoretically examines kernel regression through classical statistical estimators and random matrix theory.

In the variance-limited regime, scaling laws in the context of random feature models \citep{rahimi2008weighted, hastie2019surprises, d2020double}, deep linear models~\citep{advani2017high, advani2020high}, one hidden layer networks \citep{mei2019generalization, adlam2020neural, adlam2020understanding}, and wide neural networks treated as Gaussian processes or trained in the NTK regime \citep{lee2019wide, Dyer2020Asymptotics, andreassen2020, geiger2020scaling} have been studied. In particular, this behavior was used in \citep{kaplan2020scaling} to motivate a particular ansatz for simultaneous scaling with data and model size. The resolution-limited analysis can perhaps be viewed as an attempt to quantify the \emph{ideal-world} generalization error of \cite{nakkiran2021the}.

This work makes use of classic results connecting the spectrum of a smooth kernel to the geometry it is defined over \citep{weyl1912asymptotische, reade1983eigenvalues, kuhn1987eigenvalues, ferreira2009eigenvalues} and on the scaling of iteratively refined approximations to smooth manifolds \citep{stein1999interpolation, bickel2007local, de2011approximating}.

\section{Four scaling regimes}

Throughout this work we will be interested in how the average test loss $L(D,P)$ depends on the dataset size $D$ and the number of model parameters $P$.
Unless otherwise noted, $L$ denotes the test loss averaged over initialization of the parameters and draws of a size $D$ training set. Some of our results only pertain directly to the scaling with width $w \propto \sqrt{P}$, but we expect many of the intuitions apply more generally. We use the notation $\alpha_{D}$, $\alpha_{P}$, and $\alpha_{W}$ to indicate scaling exponents with respect to dataset size, parameter count, and width. All proofs appear in the supplement.

\subsection{Variance-limited exponents}\label{subsec:var_limited}

In the limit of large $D$ the outputs of an appropriately trained network approach a limiting form with corrections which scale as $D^{-1}$. Similarly, recent work shows that wide networks have a smooth large $P$ limit \citep{Jacot2018ntk}, where fluctuations scale as $1/\sqrt{P}$. If the loss is sufficiently smooth
%about this limiting model 
then its value will approach the asymptotic loss with corrections proportional to the variance ($1/D$ or $1/\sqrt{P}$). 
In Theorem~\ref{thm:loss_var} we present sufficient conditions on the loss to ensure this variance dominated scaling. We note, these conditions are satisfied by mean squared error and cross-entropy loss, though we conjecture the result holds even more generally.
\begin{thm}\label{thm:loss_var}
Let $\ell(f)$ be the test loss as a function of network output, ($L=\mathbb{E}\left[\ell(f)\right]$),
and let $f_{T}$ be the network output after $T$ training steps,
thought of as a random variable over weight initialization,
draws of the training dataset, and optimization seed. Further
let $f_{T}$ be concentrating with
$\mathbb{E}[\left(f_{T}-\mathbb{E}[f_{T}]\right)^{k}]=\mathcal{O}\left(\epsilon\right) \forall k\geq2$. If $\ell$ is a finite-degree polynomial, or has bounded second derivative, or is 2-H\"older, then $\mathbb{E}\left[\ell(f_{T})\right]-\ell\left(\mathbb{E}\left[f_T\right]\right)=\mathcal{O}(\epsilon)$.
\end{thm}

\subsubsection{Dataset scaling}

Consider a neural network, and its associated training loss $L_{\textrm{train}}(\theta)$.
For every value of the weights, the training loss, thought of as a random variable over draws of a training set of size $D$, concentrates around the population loss, with a variance which scales as $\mathcal{O}\left(D^{-1}\right)$. This is because if the optimization procedure is sufficiently smooth, the trained weights, network output, and higher moments, will approach their infinite $D$ values, $\mathbb{E}_{D}\left[\left(f_{T}-\mathbb{E}_{D}\left[f_{T}\right]\right)^{k}\right]=\mathcal{O}\left(D^{-1}\right)$. Here, the subscript $D$ on the expectation indicates an average over draws of the training set. This scaling together with Theorem~\ref{thm:loss_var} gives the variance limited scaling of loss with dataset size.

This concentration result with respect to dataset size has appeared for linear models in \cite{rahimi2008weighted} and for single hidden layer networks with high-dimensional input data in \cite{mei2019generalization, adlam2020neural, adlam2020understanding}. In the supplement, we prove this for GD and SGD with polynomial loss as well as present informal arguments more generally. Additionally, we present examples violating the smoothness assumption and exhibiting different scaling.

\subsubsection{Large width scaling}
We can make a very similar argument in the $w \to \infty$ limit. It has been shown that the predictions from an infinitely wide network, either under Bayesian inference \citep{nealthesis,lee2018deep}, or when trained via gradient descent \citep{Jacot2018ntk, lee2019wide}, approach a limiting distribution at large width equivalent to a linear model with random features. Furthermore, corrections to the infinite width behavior are controlled by the variance of the full model around the linear model predictions. This variance (and higher moments) have been shown to scale as $1/w$ \citep{Dyer2020Asymptotics, yaida2019non, andreassen2020}, $\mathbb{E}_{w}\left[\left(f_{T}-\mathbb{E}_{w}\left[f_{T}\right]\right)^{k}\right]=\mathcal{O}\left(w^{-1}\right)$. 
Theorem~\ref{thm:loss_var} then implies the loss will differ from its $w=\infty$ limit by a term proportional to $1/w$.

We note that there has also been work studying the combined large depth and large width limit, where \cite{Hanin2020Finite} found a well-defined infinite size limit with controlled fluctuations in randomly initialized deep neural networks. In any such context where the trained model predictions concentrate, we expect the loss to scale with the variance of the model output. In the case of linear models, studied below, the variance is $\cO(P^{-1})$ rather than $\cO(\sqrt{P})$, and we see the associated variance scaling in this case.

\subsection{Resolution-limited exponents}
\label{subsec:resolution-limited}

In this section we consider training and test data drawn uniformly from a compact $d$-dimensional manifold, $x\in\mathcal{M}_{d}$, and targets given by some smooth function $y=\mathcal{F}(x)$ on this manifold.\\

\subsubsection{Overparameterized dataset scaling}

Consider the double limit of an \op{} model with large training set size, $P \gg D \gg 1$. We further consider \emph{well-trained} models, i.e. models that interpolate all training data. The goal is to understand $L(D)$.
If we assume that the learned model $f$ is sufficiently smooth, then the dependence of the loss on $D$ can be bounded in terms of the dimension of the  data manifold $\mathcal{M}_{d}$.

Informally, if our train and test data are drawn i.i.d. from the same manifold, then the distance from a test point to the closest training data point decreases as we add more and more training data points. In particular, this distance scales as $\mathcal{O}(D^{-1/d})$ \citep{levina2005maximum}. Furthermore, if $f$, $\mathcal{F}$ are both sufficiently smooth, they cannot differ too much over this distance. If in addition the loss function, $L$, is a smooth function vanishing when $f=\mathcal{F}$, we have $L=\mathcal{O}(D^{-1/d})$. This is summarized in the following theorem.
\begin{thm}\label{thm:interp_data}
Let $L(f)$, $f$ and $\mathcal{F}$ be Lipschitz with constants $K_{L}$, $K_{f}$, and $K_{\mathcal{F}}$. Further let $\mathcal{D}$ be a training dataset of size $D$ sampled i.i.d from $\mathcal{M}_{d}$ and let $f(x)=\mathcal{F}(x),\,\, \forall x \in\mathcal{D}$
then $L(D)=\cO\left(K_{L}\textrm{max}{(K_{f},K_{\mathcal{F}})}D^{-1/d}\right)$.
\end{thm}

\subsubsection{Underparameterized parameter scaling}

We will again assume that $\mathcal{F}$ varies smoothly on an underlying compact $d$-dimensional manifold $\cM_d$.  We can obtain a bound on $L(P)$ if we imagine that $f$ approximates $\mathcal{F}$ as a piecewise function with roughly $P$ regions (see \cite{sharma2020neural}).
Here, we instead make use of the argument from the over-parameterized, resolution-limited regime above. If we construct a sufficiently smooth estimator for $\mathcal{F}$ by interpolating among $P$ randomly chosen points from the (arbitrarily large) training set, then by the argument above the loss will be bounded by $\cO(P^{-1/d})$.  

\begin{thm}\label{thm:interp_params}
Let $L(f)$, $f$ and $\mathcal{F}$ be Lipschitz with constants $K_{L}$, $K_{f}$, and $K_{\mathcal{F}}$. Further let $f(x)=\mathcal{F}(x)$ for $P$ points sampled i.i.d from $\mathcal{M}_{d}$
then $L(P)=\cO\left(K_{L}\textrm{max}{(K_{f},K_{\mathcal{F}})}P^{-1/d}\right)$.
\end{thm}

\subsubsection{From bounds to estimates}

Theorems~\ref{thm:interp_data} and \ref{thm:interp_params} are phrased as bounds, but we expect the stronger statement that these bounds  also generically serve as estimates, so that eg $L(D)=\Omega(D^{-c/d})$ for $c\geq 2$, and similarly for parameter scaling. If we assume that $\mathcal{F}$ and $f$ are analytic functions on $\mathcal{M}_{d}$ and that the loss function $L(f,\mathcal{F})$ is analytic in $f-\mathcal{F}$ and minimized at $f=\mathcal{F}$, then the loss at a given test input, $x_{\textrm{test}}$, can be expanded around the nearest training point, $\hat{x}_{\textrm{train}}$,
$
L(x_{\textrm{test}})=\sum_{m=n \geq 2}^{\infty}a_{m}(\hat x_{\textrm{train}})(x_{\textrm{test}}-\hat{x}_{\textrm{train}})^{m}
$,\footnote{For simplicity we have used a very compressed notation for multi-tensor contractions in higher order terms.} 
where the first term is of finite order $n \geq 2$ because the loss vanishes at the training point. As the typical distance between nearest neighbor points scales as $D^{-1/d}$ on a $d$-dimensional manifold (an observation also made in \cite{spigler2020asymptotic}), the loss will be dominated by the leading term, $L \propto D^{-n/d}$, at large $D$.  Note that if the model provides an accurate piecewise linear approximation, we will generically find $n\geq 4$.

\begin{figure*}[t!]
\centering
     \begin{subfigure}
     \centering
         \includegraphics[width=1.0\textwidth]{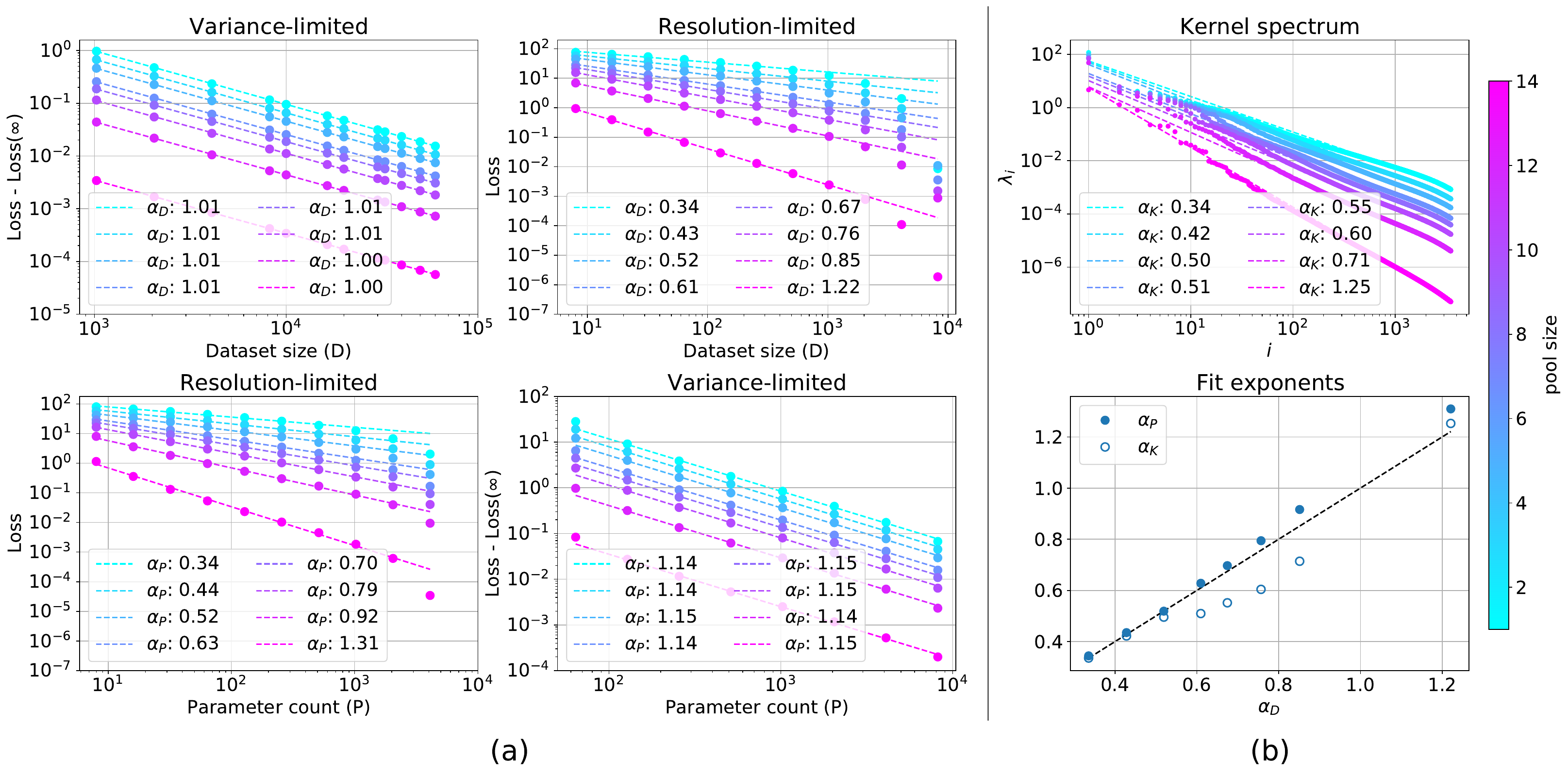}
     \end{subfigure}

    \vspace{-0.2cm}
        \caption{\small\textbf{(a) Random feature models exhibit all four scaling regimes.} Here we consider linear teacher-student models with random features trained with MSE loss to convergence. We see both \emph{variance-limited} scaling \textbf{(top-left, bottom-right)} and \emph{resolution-limited} scaling \textbf{(top-right, bottom-left)}. Data is varied by downsampling MNIST by the specified pool size.
        \textbf{(b) Duality and spectra in random feature models.} Here we show the relation between the decay of the kernel spectra, $\alpha_{K}$, and the scaling of the loss with number of data points, $\alpha_{D}$, and with number of parameters, $\alpha_{P}$. \textbf{(top)} The spectra of kernels derived from random fully-connected deep neural networks on pooled MNIST \textbf{(bottom)} appear well described by a power-law decay. The theoretical relation $\alpha_{D}=\alpha_{P}=\alpha_{K}$ is given by the dashed black line. 
        }
        \label{fig:kernel_banner}
        \label{fig:kernel_exponents}
\end{figure*}

\subsection{Explicit realization in linear random feature models}\label{sec:kernel}

In the proceeding sections we have conjectured typical case scaling relations for a model's test loss. We have further given intuitive arguments for this behavior which relied on smoothness assumptions on the loss and training procedure. In this section, we provide a concrete realization of all four scaling regimes within the context of linear models constructed from random features. Of particular interest is the resolution-limited regime, where the scaling of the loss is a consequence of the linear model kernel spectrum -- the scaling of \op{} models with dataset size and \up{} models with parameters is a consequence of a classic result, originally due to~\cite{weyl1912asymptotische}, bounding the spectrum of sufficiently smooth kernel functions by the dimension of the manifold they act on.

Linear predictors serve as a model system for learning.
Such models are used frequently in practice when more expressive models are unnecessary or infeasible \citep{mccullagh1989generalized, rifkin2007notes, hastie2009elements} and also serve as an instructive test bed to study training dynamics~\citep{advani2020high, hastie2019surprises, nakkiran2019more}.
Furthermore, in the large width limit, deep neural networks behave as Gaussian processes \citep{nealthesis, lee2018deep, matthews2018, novak2018bayesian, garriga2018deep, yang2019scaling} and in the low-learning rate regime of gradient-descent optimization \citep{lee2019wide, lewkowycz2020large, huang_largelr}, deep neural networks behave as a particular class of linear models \citep{Jacot2018ntk, lee2019wide, chizat2019lazy}. Hence, linear predictors constructed from random features provide an accurate description of deep neural networks in the large width limit.

Here we discuss linear models in general terms, though the results immediately hold for the special cases of  wide, deep neural networks. We focus on teacher-student models, in which the teacher generates samples from which the student model learns. We will assume student weights initialized to zero and trained with mean squared error (MSE) loss to their global optimum.

We consider a linear teacher $F$ and student $f$,
\begin{align}
    F(x)&=\sum_{M=1}^{S}\omega_{M}F_{M}(x) & f(x)&=\sum_{\mu=1}^{P}\theta_{\mu}f_{\mu}(x) \,.
\end{align}

\noindent Here $\{F_{M}\}$ are a (potentially infinite) collection of features. The teacher weights, $\omega_{M}$, are sampled from a Normal distribution $\omega\sim\mathcal{N}(0,1/S)$ and are averaged over in the test loss. The student has learnable parameters $\{ \theta_{\mu} \}$ and is built out of a subset of the teacher features. To vary the number of parameters in this simple model, we construct $P$ features, $f_{\mu=1,\ldots,P}$, by introducing a projector $\cP$ onto a $P$-dimensional subspace of the teacher features, $f_{\mu}=\sum_{M}\mathcal{P}_{\mu M}F_{M}$. We train by sampling a training set of size $D$ and minimizing the MSE loss, $L_{\textrm{train}}=\frac{1}{2D}\sum_{a=1}^{D}\left(f(x_{a})-F(x_{a})\right)^{2}$. We are interested in the test loss averaged over draws of our teacher weights and training dataset.

The infinite data test loss, $L(P):=\lim_{D\rightarrow\infty}L(D,P)$, takes the form,
\es{eq:up_infd}{
L(P)&=\frac{1}{2 S}\textrm{Tr}\left[\mathcal{C}-\mathcal{C}\cP^{T}\left(\cP\mathcal{C}\cP^{T}\right)^{-1}\cP\mathcal{C} \right]\,.
}
Here we have introduced the feature-feature second moment-matrix, $\mathcal{C}=\lexpp{x}F(x)F^{T}(x)\rexp$. If the teacher and student features had the same span, this would vanish, but due to the mismatch the loss is nonzero. 

On the other hand, if we keep a finite number of training points, but allow the student to use all of the teacher features, the test loss, $L(D):=\lim_{P\rightarrow S}L(D,P)$, takes the form, 
\es{eq:op_infp}{
L(D) = \frac{1}{2}\lexpp{x}\mathcal{K}(x,x)-\vec{\mathcal{K}}(x)\bar{\mathcal{K}}^{-1}\vec{\mathcal{K}}(x)\rexp\,.
}
Here, $\mathcal{K}(x,x')$ is the data-data second moment matrix, $\vec{\mathcal{K}}$ indicates restricting one argument to the $D$ training points, while $\bar{\mathcal{K}}$ indicates restricting both. This test loss vanishes as the number of training points becomes infinite but is non-zero for finite training size. 

We present a full derivation of these expressions in the supplement. In the remainder of this section, we explore the scaling of the test loss with dataset and model size.

\subsubsection{Variance-limited scaling}

To derive the limiting expressions \eqref{eq:up_infd} and \eqref{eq:op_infp} for the loss one makes use of the fact that the sample expectation of the second moment matrix over the finite dataset, as well as the finite feature set, is close to the full covariance,
$\frac{1}{D}\sum_{a=1}^{D}F(x_{a})F^{T}(x_{a})\,=\,\mathcal{C}+\delta\mathcal{C}$, $\frac{1}{P}f^{T}(x)f(x'),=\,\mathcal{K}+\delta\mathcal{K}$, with the fluctuations satisfying $\lexpp{D}\delta C^{2}\rexp=\mathcal{O}(D^{-1})$ and $\lexpp{P}\delta K^{2}\rexp=\mathcal{O}(P^{-1})$, where expectations are taken over draws of a dataset of size $D$ and over feature sets. Using these expansions yields the variance-limited scaling, $L(D,P)-L(P)=\mathcal{O}(D^{-1})$, $L(D,P)-L(D)=\mathcal{O}(P^{-1})$ in the \up{} and \op{} settings, respectively.

In Figure~\ref{fig:kernel_banner}a we see evidence of these scaling relations for features built from randomly initialized ReLU deep neural networks on coarse-grained versions of the MNIST dataset obtained by local averaging over the images. We see that in the variance-limited regimes the scaling exponent is independent of the modification to the training data. In the supplement, we provide an in-depth derivation of this behavior and expressions for the leading contributions to $L(D,P)-L(P)$ and $L(D,P)-L(D)$.

\subsubsection{Resolution-limited scaling}

We now would like to analyze the scaling behavior of our linear model in the resolution-limited regimes, that is the scaling with $P$ when $1 \ll P \ll D$ and the scaling with $D$ when $1 \ll D \ll P$. In these cases, the scaling is controlled by the shared spectrum of $\mathcal{C}$ or $\mathcal{K}$. This spectrum is often well described by a power-law, where eigenvalues $\lambda_{i}$ satisfy $\lambda_{i}=\frac{1}{i^{1+\alpha_{K}}}$.
See Figure~\ref{fig:kernel_exponents}b for example spectra on pooled MNIST.

In this case, we will argue that the losses also obey a power law scaling, with the exponents controlled by the spectral decay factor, $1+\alpha_{K}$,
\es{eq:ideal_scaling}{
L(D)\propto D^{-\alpha_{K}}\,, \ \ \ L(P)\propto P^{-\alpha_{K}}\,.
}
In other words, in this setting, $\alpha_{P}=\alpha_{D}=\alpha_{K}$. This is supported empirically in Figure~\ref{fig:kernel_exponents}b. For other derivations of dataset scaling for kernel regression, see \cite{spigler2020asymptotic, bordelon2020spectrum}. We then argue that when the kernel function $\mathcal{K}$ is sufficiently smooth on a manifold of dimension $d$, $\alpha_{K}\propto d^{-1}$, thus realizing the more general resolution-limited picture described above.

\subsubsection{From spectra to scaling laws for the loss}
To be concrete let us focus on the \op{} loss. If we introduce the notation $e_{i}$ for the eigenvectors of $\mathcal{C}$ and $\bar{e}_{i}$ for the eigenvectors of $\frac{1}{D}\sum_{a=1}^{D}F(x_a)F^{T}(x_{a})$, the loss becomes,
\es{eq:op_eigen_loss}{
L(D)=\frac{1}{2}\sum_{i=1}^{S}\lambda_{i}(1-\sum_{j=1}^{D}(e_{i}\cdot\bar{e}_{j})^{2})\,.
}
Before discussing the general asymptotic behavior of \eqref{eq:op_eigen_loss}, we can gain some intuition by considering the case of large $\alpha_{K}$. In this case, $\bar{e}_{j}\approx e_{j}$ (see e.g. \cite{loukas2017close}), we can simplify \eqref{eq:op_eigen_loss} to,
\es{eq:up_eigen_loss_simp}{
L(D)&\propto\sum_{i=D+1}^{\infty}\frac{1}{i^{1+\alpha_{K}}}=%\zeta(\alpha+1,D+1)\\
\alpha_{K}D^{-\alpha_{K}}+\mathcal{O}(D^{-\alpha_{K}-1})\,.
}
More generally in the supplement, following \cite{bordelon2020spectrum, canatar2020statistical} we use replica theory methods to derive $L(D)\propto D^{-\alpha_{K}}$ and $L(P)\propto P^{-\alpha_{K}}$, without requiring the large $\alpha_{K}$ limit.

\subsubsection{Data manifolds and kernels}
In Section \ref{subsec:resolution-limited}, we discussed a simple argument that resolution-limited exponents $\alpha \propto 1/d$, where $d$ is the dimension of the data manifold.  Our goal now is to explain how this connects with the linearized models and kernels discussed above: how does the spectrum of eigenvalues of a kernel relate to the dimension of the data manifold?

The key point is that sufficiently \emph{smooth} kernels must have an eigenvalue spectrum with a bounded tail. Specifically, a $C^t$ kernel on a $d$-dimensional space must have  eigenvalues $\lambda_n \lesssim \frac{1}{n^{1 + t/d}}$ \citep{kuhn1987eigenvalues}.  In the generic case where the covariance matrices we have discussed  can be interpreted as kernels on a manifold, and they have spectra \emph{saturating} the bound, linearized models will inherit scaling exponents given by the dimension of the manifold.

As a simple example, consider a $d$-torus.  In this case we can study the Fourier series decomposition, and examine the case of a kernel $K(x-y)$.  This must  take the form $K =  \sum_{n_I}\left[a_{n_I} \sin(n_I \cdot (x - y) ) + b_{n_I} \cos(n_I \cdot (x - y) )\right]\nonumber$,
where $n_I = (n_1, \cdots, n_d)$ are integer indices, and $a_{n_I}$, $b_{n_I}$ are the overall Fourier coefficients.  To guarantee that $K$ is a $C^t$ function, we must have $a_{n_I}, b_{n_I} \lesssim \frac{1}{n^{d + t}}$ where $n^d = N$ indexes the number of $a_{n_I}$ in decreasing order.  But this means that in this simple case, the tail eigenvalues of the kernel must be bounded by $\frac{1}{N^{1+ t/d}}$ as $N \to \infty$.

\begin{figure*}[t]
     \centering
         \includegraphics[width=1.0\textwidth]{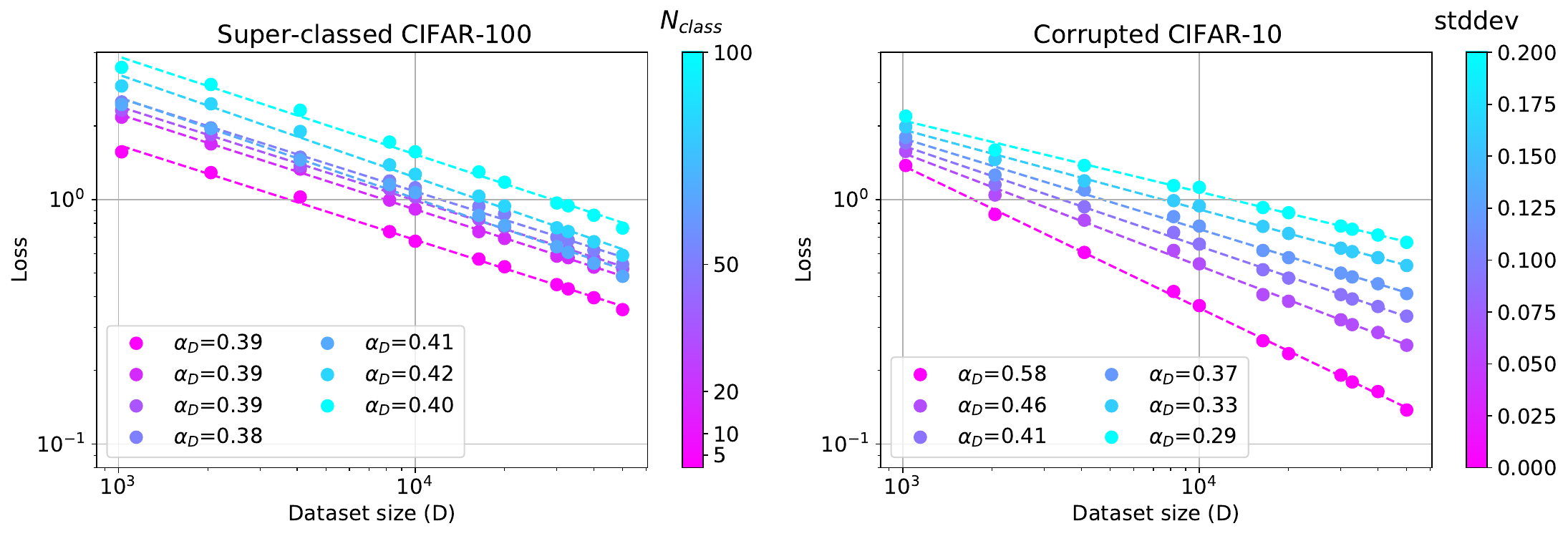}
     %\hfill
        %\vspace{-0.5cm}
        \caption{\small\textbf{Effect of data distribution on scaling exponents.} For CIFAR-100 superclassed to $N$ classes \textbf{(left)}, we find that the number of target classes does not have a visible effect on the scaling exponent. \textbf{(right)} For CIFAR-10 with the addition of Gaussian noise to inputs, we find the strength of the noise has a strong effect on performance scaling with dataset size. All models are WRN-28-10.
        }
        \label{fig:data_distribution}
        \label{fig:cifar_data_dep}
\end{figure*}

\subsection{Duality}
We argued above that for kernels with pure power-law spectra, the asymptotic scaling of the \up{} loss with respect to model size and the \op{} loss with respect to dataset size share a common exponent. In the linear setup at hand, the relation between the \up{} parameter dependence and \op{} dataset dependence is even stronger. The \up{} and \op{} losses are directly related by exchanging the projection onto random features with the projection onto random training points. 
Note, sample-wise double descent observed in~\cite{nakkiran2019more} is a concrete realization of this duality for a simple data distribution.
In the supplement, we present examples exhibiting the duality of the loss dependence on model and dataset size outside of the asymptotic regime.

\section{Experiments}

\subsection{Deep teacher-student models}
Our theory can be tested very directly in the teacher-student framework, in which a \emph{teacher} deep neural network generates synthetic data used to train a \emph{student} network.  Here, it is possible to generate unlimited training samples and, crucially, controllably tune the dimension of the data manifold. 
We accomplish the latter by scanning over the dimension of the inputs to the teacher.  We have found that when scanning over both model size and dataset size, the interpolation exponents closely match the prediction of $4/d$. The dataset size scaling is shown in Figure~\ref{fig:data_manifold_intro}, while model size scaling experiments appear in the supplement and have previously been observed in \cite{sharma2020neural}.

\subsection{Variance-limited scaling in the wild}

Variance-limited scaling (\ref{subsec:var_limited}) can be universally observed in real datasets.
Figure~\ref{fig:four_regimes}a (top-left, bottom-right) measures the variance-limited dataset scaling exponent $\alpha_D$ and width scaling exponent $\alpha_W$. In both cases, we find striking agreement with the theoretically predicted values $\alpha_{D}, \alpha_{W} = 1$ across a variety of dataset, neural network architecture, batch size in stochastic gradient descent, and loss type. Our testbed includes deep fully-connected and convolutional networks with ReLU or Erf nonlinearities and MSE or cross-entropy losses. The supplement contains experimental details.

\subsection{Resolution-limited scaling in the wild}

In addition to teacher-student models, we explored resolution-limited scaling behavior in the context of standard classification datasets. 
Wide ResNet (WRN) models~\citep{zagoruyko2016wide} were trained for a fixed number of steps with cosine decay.
In Figure~\ref{fig:visualizing_interpolation}b we also include data from a four hidden layer convolutional neural network (CNN) detailed in the supplement. As detailed above, we find dataset dependent scaling behavior in this context.

We further investigated the effect of the data distribution on the resolution-limited exponent, $\alpha_{D}$, by tuning the number of target classes and input noise (Figure~\ref{fig:cifar_data_dep}).
To probe the effect of the number of classes, we constructed tasks derived from CIFAR-100 by grouping classes into broader semantic categories. We found that performance depends on the number of categories, but $\alpha_{D}$ is insensitive to this number. In contrast, the addition of Gaussian noise had a more pronounced effect on $\alpha_{D}$. This suggest a picture in which the neural network learns to model the input data manifold, independent of the classification task, consistent with observations in \cite{nakkiran2020distributional, Grathwohl2020Your}. 

We also explored the effect of aspect ratio on dataset scaling, finding that the exponent magnitude increases with width up to a critical width, while the dependence on depth is milder (see supplement).

\section{Discussion}\label{sec:Discussion}

We have presented a framework for categorizing neural network scaling laws, along with derivations that help to explain their origins.
Crucially, our predictions agree with empirical findings in settings which have often proven challenging for theory -- deep neural networks on real datasets. The variance-scaling regime yields, for smooth test losses, a universal prediction of $\alpha_{D} = 1$ (for $D \gg P$) and $ \alpha_{W} = 1$ (for $w \gg D$). The resolution-limited regime yields exponents whose numerical value is variable and data and model dependent. 

There are a number of intriguing directions for future work. The invariance of the dataset scaling exponent to superclassing (Figure~\ref{fig:cifar_data_dep}) suggests that deep networks may be largely learning properties of the input data manifold -- akin to unsupervised learning -- rather than significant task-specific structure, which may shed light on the versatility of learned deep network representations for different downstream tasks. Another direction for future research is to more explicitly derive within the theory the effects of ``feature learning." While the random feature linear models we have discussed are in exact correspondence with deep neural networks in the large-width limit and have been a useful theoretical testbed across a variety of problems, the kernels associated with networks of finite-depth and finite-width evolve dynamically during the course of training.

\subsection{Limitations} One limitation is that our theoretical results are asymptotic, while experiments are performed with finite models and datasets. This is apparent in the resolution-limited regime which requires a hierarchy ($D\gg P$ or $P\gg D$). In Figures~\ref{fig:four_regimes}a and \ref{fig:kernel_banner}a top-right (bottom-left), we see a breakdown of the predicted scaling behavior as $D$ $(P)$ become large and the hierarchy is lost. Furthermore, in the resolution-limited regime for deep networks, our theoretical tools rely on positing the existence of a data manifold. A precise definition of the data manifold, however, is lacking, forcing us to use imperfect proxies, such as the nearest-neighbor distances of final embedding layers from a trained network. \\

\section{Outlook}

Modern deep learning, in the era of large datasets, models, and computational power, has often made progress through extensive amounts of experimentation and trial-and-error. A theoretical understanding of deep learning that is grounded in experiments and strives to bridge the gap between mathematically rigorous theory on the one hand, and realistic settings on the other, could be scientifically important for guiding the field. Our treatment of neural scaling laws in this work touches on classic aspects of generalization within learning theory but derives new results through realistic data assumptions and identifying and deriving a taxonomy for scaling regimes. Our approach is guided by the theoretical simplicity, realistic modeling, and experimental verification that is characteristic of theory construction in physics; we also leverage results from statistical physics approaches to deep learning in our derivations.

Looking further afield, it is an interesting question whether qualitatively new behavior can emerge in large neural models trained on rich datasets, or whether such models are natural extensions of smaller-scale models. The exploration of so-called emergent abilities within neural-based language models is an active area of research. Further investigation into these questions through the theoretical methods and scientific approaches of physics -- calling for a realistic modeling of data and neural representations -- may help shed light on our understanding of learning in deep neural networks. \\

\noindent \textbf{Acknowledgements.} The authors would like to thank
Guy Gur-Ari,
Boris Hanin,
Tom Henighan,
Danny Hernandez,
Aitor Lewkowycz,
Sam McCandlish,
Preetum Nakkiran,
Behnam Neyshabur,
Jeffrey Pennington,
Vinay Ramasesh,
Dan Roberts,
Jonathan Rosenfeld,
Jascha Sohl-Dickstein,
and
Lechao Xiao
for conversations during the completion of this work. US completed a portion of this work during an internship at Google. JK and US were supported in part by Open Philanthropy.\\

\bibliography{references}
\bibliographystyle{unsrtnat}

\normalsize
\onecolumn
\clearpage
\appendix
\label{sec:curr_appendix}

\begin{center}
\textbf{\Large Supplemental Material}
\end{center}
\label{}
\setcounter{equation}{0}
\setcounter{figure}{0}
\setcounter{table}{0}
\setcounter{page}{1}
\setcounter{section}{0}
\makeatletter
\renewcommand{\theequation}{S\arabic{equation}}
\renewcommand{\thefigure}{S\arabic{figure}}
\renewcommand{\bibnumfmt}[1]{[S#1]}

\setcounter{page}{17}

\section{Experimental setup}
\label{supp:experiment setup}

\subsection{Figure 1 (top-left)}

Experiments utilize relatively small models, with the number of trainable parameteters $P \sim \mathcal{O}(1000)$, trained with full-batch gradient descent (GD) and small learning rate on datasets of size $D \gg P$. Each data point in the figure represents an average over subsets of size $D$ sampled from the full dataset. Experiments are done using the Neural Tangents~\citep{neuraltangents2020} library based on JAX~\citep{jaxrepro}.
 All experiment except denoted as (CNN), use 3-layer, width-8 fully-connected networks. Convolutional neural network (CNN) architecture used is Myrtle-5 network \citep{Shankar2020NeuralKW} with 8 channels. ReLU activation function with critical initialization~\citep{schoenholz2016deep, lee2018deep, xiao18a} was used. Unless specified cross-entropy loss was used. We performed full-batch gradient descent update for all dataset sizes without $L_2$ regularization. 20 different training data sampling seeds were averaged for each point. For fully-connected networks, input pooling of size 4 was performed for CIFAR-10/100 dataset and pooling of size 2 was performed for MNIST and Fashion-MNIST dataset. This was to reduce number of parameters in the input layer (\# of pixels $\times$ width) which can be quite large even for small width networks. 
 
\subsection{Figure 1 (top-right)}
\begin{figure*} %[h!]
     \centering
     \begin{subfigure}
         \centering
         \includegraphics[width=0.8\textwidth]{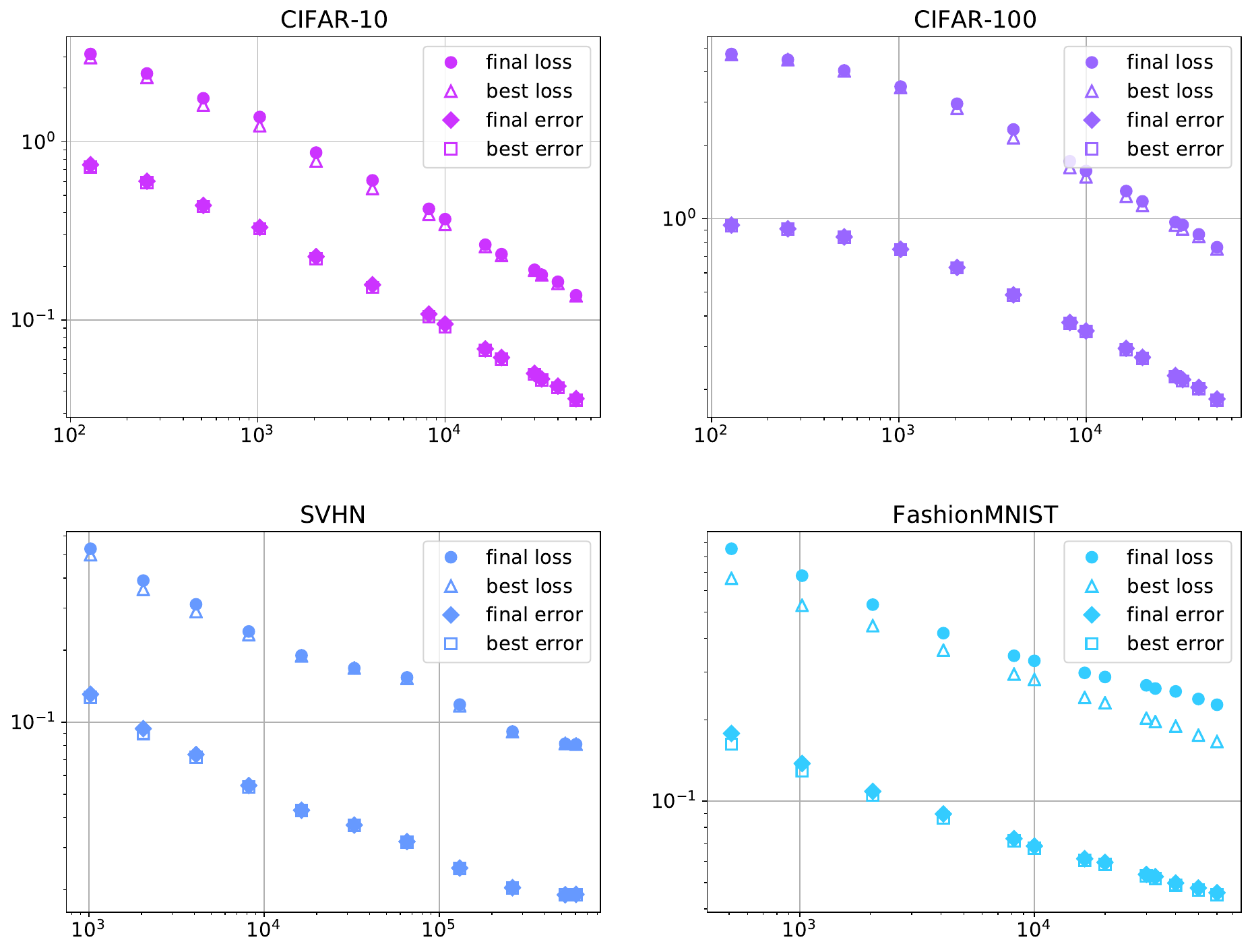}
         %\caption{}
     \end{subfigure}
     \hfill
        \caption{\small\textbf{Alternate metrics and stopping conditions.} We find similar scaling behavior for both the loss and error, as well as for final and best (early stopped) metrics.}
        \label{fig:alt_metrics}
\end{figure*}

All experiments were performed using a Flax \citep{flax2020github} implementation of Wide ResNet 28-10 \citep{zagoruyko2016wide}. Models were trained for 78125 total steps with a cosine learning rate decay \citep{loshchilov2016sgdr} and an augmentation policy consisting of random flips and crops. We report final loss, though we found no qualitative difference between using final loss, best loss, final accuracy or best accuracy (see Figure~\ref{fig:alt_metrics}).

\subsection{Figure 1 (bottom-left)}
The setup was identical to Figure 1 (top-right) except that the model considered was a depth 10 residual network with varying width. 
 
\subsection{Figure 1 (bottom-right)} 
 
 Experiments are done using Neural Tangents. All experiments use 100 training samples and two hidden layer fully-connected networks of varying width (ranging from $w = 64$ to $w=11,585$) with ReLU nonlinearities unless specified as Erf. Full-batch gradient descent and cross-entropy loss were used unless specified as MSE, and the figure shows curves from a random assortment of training times ranging from 100 to 500 steps (equivalently, epochs). Training was done with learning rates small enough so as to avoid catapult dynamics \citep{lewkowycz2020large} and no $L_2$ regularization; in such a setting, the infinite-width learning dynamics is known to be equivalent to that of linearized models \citep{lee2019wide}. Consequently, for each random initialization of the parameters, the test loss of the finite-width linearized model was additionally computed in the identical training setting. This value approximates the limiting behavior $L(\infty)$ known theoretically and is subtracted off from the final test loss of the (nonlinear) neural network before averaging over 50 random initializations to yield each of the individual data points in the figure.  

\subsection{Deep teacher-student models}

\begin{figure*} %[t]
     \centering
     \begin{subfigure}
         \centering
         \includegraphics[width=0.47\textwidth]{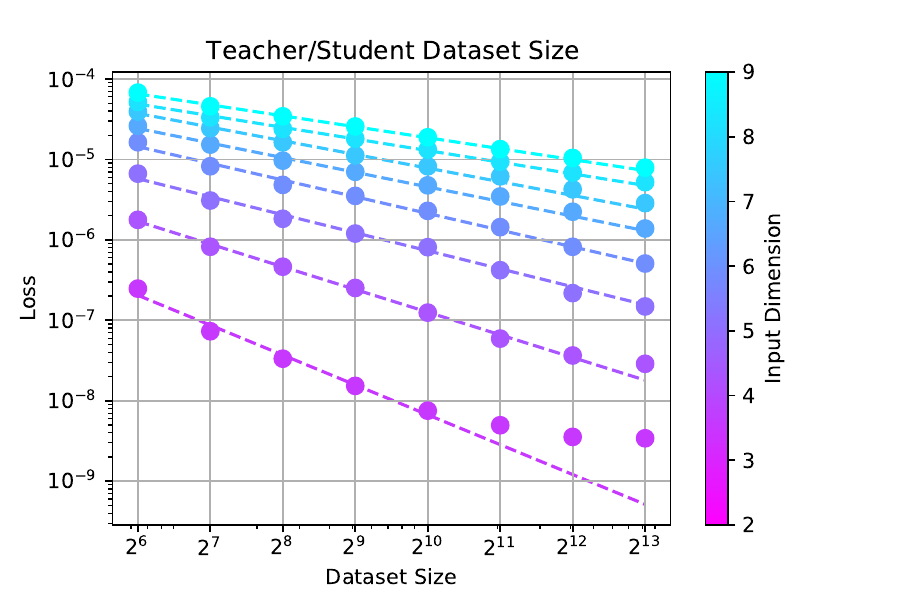}
     \end{subfigure}
        \caption{\small This figure shows scaling trends of MSE loss with dataset size for teacher/student models. The exponents extracted from these fits and their associated input-space dimensionalities are shown in Figure 1b.}
        %\ref{fig:data_manifold_intro}.} 
        \label{fig:TS_banner}
\end{figure*}

The teacher-student scaling with dataset size (figure \ref{fig:TS_banner}) was performed with fully-connected teacher and student networks with two hidden layers and widths 96 and 192, respectively, using PyTorch \citep{paszke2019pytorch}.  The inputs were random vectors sampled uniformly from a hypercube of dimension $d=2,3, \cdots, 9$.  To mitigate noise, we ran the experiment on eight different random seeds, fixing the random seed for the teacher and student as we scanned over dataset sizes.  We also used a fixed test dataset, and a fixed training set, which was subsampled for the experiments with smaller $D$.  The student networks were trained using MSE loss and Adam optimizer with a maximum learning rate of $3\times 10^{-3}$, a cosine learning rate decay, and a batch size of $64$, and $40,000$ steps of training.   The test losses were measured with early stopping.  We combine test losses from different random seeds by averaging the logarithm of the loss from each seed.

In our experiments, we always use inputs that are uniformly sampled from  a $d$-dimensional hypercube, following the setup of \cite{sharma2020neural}. They also utilized several intrinsic dimension (ID) estimation methods and found the estimates were close to the input dimension, so we simply use the latter for comparisons.  
For the dataset size scans, we used randomly initialized teachers with width 96 and students with width 192.  We found similar results with other network sizes.  

The final scaling exponents and input dimensions are show in the bottom of Figure 1b.

We used the same experiments for the top of that figure, interpolating the behavior of both teacher and a set of students between two fixed training points.  The students only differed by the size of their training sets but had the same random seeds and were trained in the same way.  In that figure, the input space dimension was four.

Finally, we also used a similar setup to study variance-limited exponents and scaling.  In that case we used much smaller models, with 16-dimensional hidden layers, and a correspondingly larger learning rate.  We then studied scaling with $D$ again, with results pictured in Figure 1a.

\begin{table}[t!]
\centering
\caption{\label{tab:ResLimitedScalingCNN} \small CNN architectures for CIFAR-10, MNIST, Fashion MNIST (left), CIFAR-100 (center) and SVHN (right).}
\begin{tabular}{| c | c | c |}
\hline
 \textbf{Layer} & \textbf{Width} \\ \hline
 CNN window (3, 3) & 50 \\ \hline
 2D Max Pooling (2, 2) & \\  \hline
 CNN window (3, 3) & 100 \\  \hline
 2D Max Pooling (2, 2) & \\ \hline
 CNN window (3, 3) & 100 \\  \hline
 Dense & 64\\ \hline
 Dense & 10\\ \hline
\end{tabular}
\quad
\begin{tabular}{| c | c | c |}
\hline
 \textbf{Layer} & \textbf{Width} \\ \hline
 CNN window (3, 3) & 50 \\ \hline
 2D Max Pooling (2, 2) & \\  \hline
 CNN window (3,3) & 100 \\  \hline
 2D Max Pooling (2, 2) & \\ \hline
 CNN window (3, 3) & 200 \\  \hline
 Dense & 256\\ \hline
 Dense & 100\\ \hline
\end{tabular}
\quad
\begin{tabular}{| c | c | c |}
\hline
 \textbf{Layer} & \textbf{Width} \\ \hline
 CNN window (3, 3) & 64 \\ \hline
 2D Max Pooling (2, 2) & \\  \hline
 CNN window (3, 3) & 64 \\  \hline
 2D Max Pooling (2, 2) & \\ \hline
 Dense & 128\\ \hline
 Dense & 10\\ \hline
\end{tabular}
\end{table}

\subsection{CNN architecture for resolution-limited scaling}

\begin{figure}
    \centering
    \includegraphics[scale=0.37]{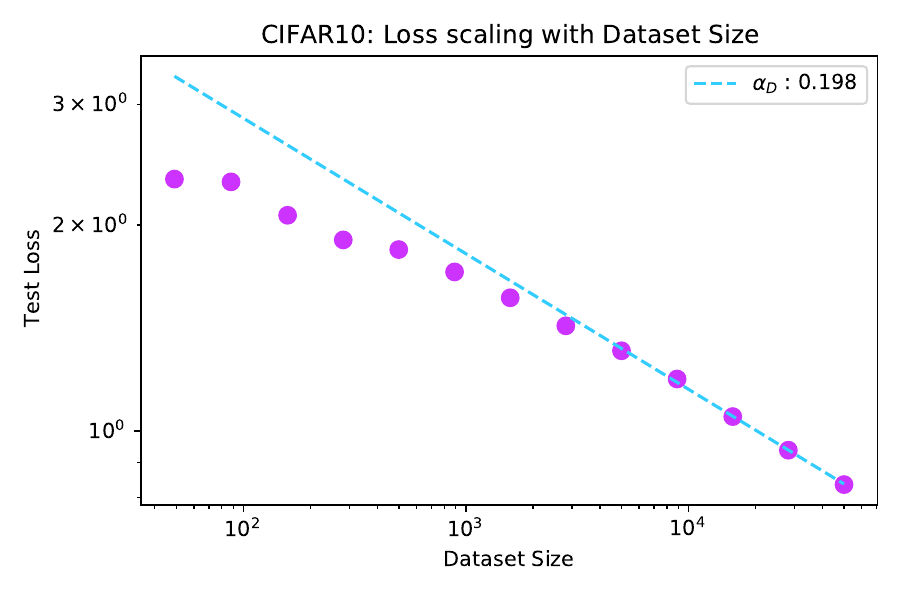}
    \includegraphics[scale=0.37]{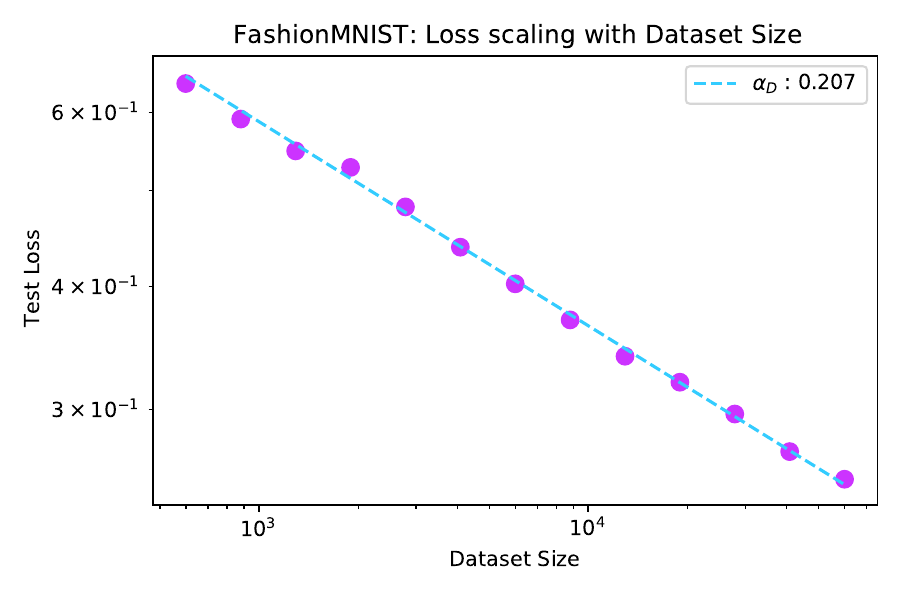}
    \includegraphics[scale=0.37]{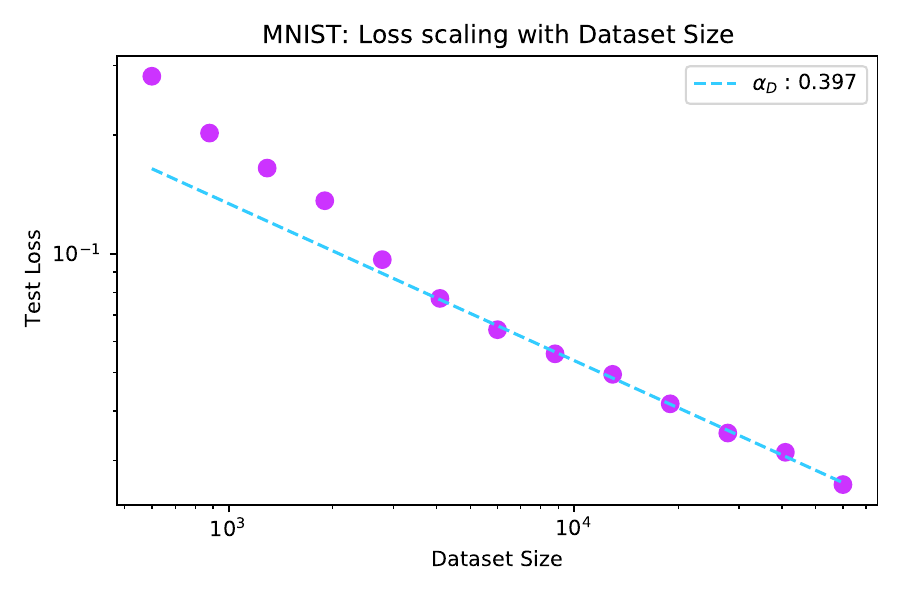}
    \includegraphics[scale=0.37]{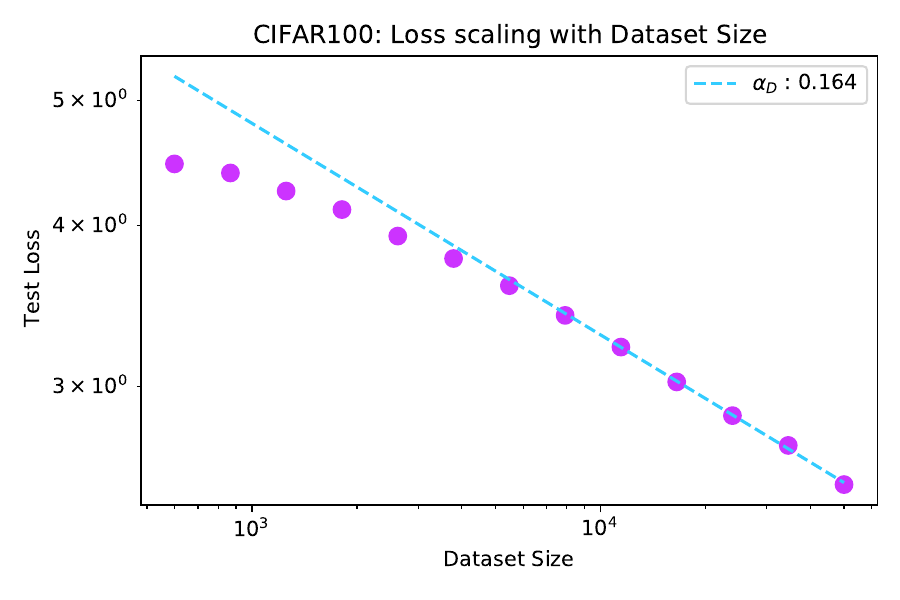}
    \includegraphics[scale=0.37]{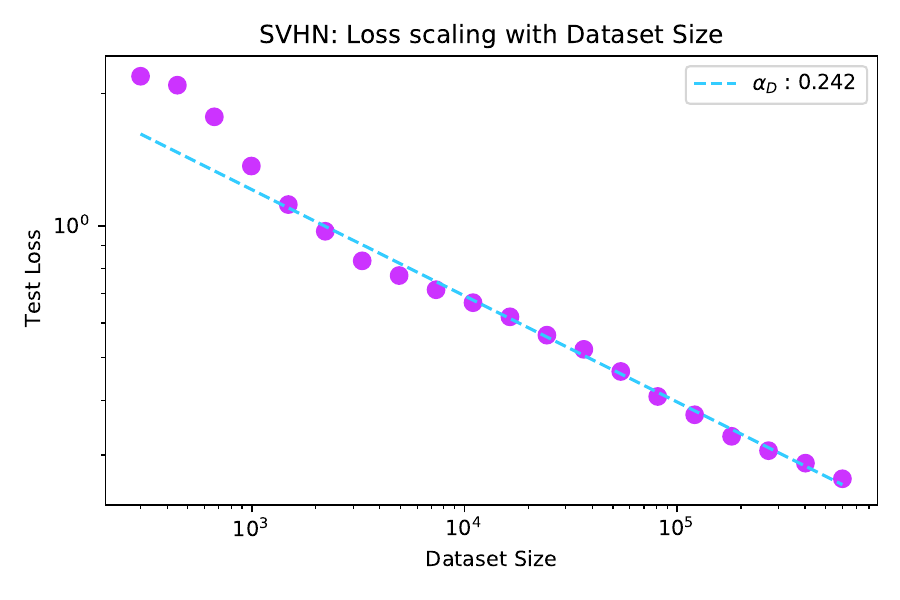}
    \caption{\small This figure shows scaling trends of cross-entropy loss with dataset size for various image datasets. The exponents extracted from these fits and their associated input-space dimensionalities are shown in Figure 1b.}
    %\ref{fig:data_manifold_intro}.}
    \label{fig:CNNsForFig2}
\end{figure}

Figure 1b
%~\ref{fig:visualizing_interpolation}b 
includes data from CNN architectures trained on image datasets. The architectures are summarized in Table \ref{tab:ResLimitedScalingCNN}. We used Adam optimizer for training with cross-entropy loss. Each network was trained for long enough to achieve either a clear minimum or a plateau in test loss. Specifically, CIFAR-10, MNIST and Fashion MNIST were trained for $50$ epochs, CIFAR-100 was trained for $100$ epochs, and SVHN was trained for $10$ epochs. The default Keras training parameters were used. In case of SVHN, we included the additional images as training data. We averaged (in log space) over $20$ runs for CIFAR-100 and CIFAR-10, $16$ runs for MNIST, $12$ runs for Fashion MNIST, and $5$ runs for SVHN. The results of these experiments are shown in Figure \ref{fig:CNNsForFig2}.

The measurement of input-space dimensionality for these experiments was done using the nearest-neighbour algorithm, described in detail in Appendix B and C in \cite{sharma2020neural}. We used 2, 3 and 4 nearest neighbors and averaged over the three.

\subsection{Teacher-student experiment for scaling of loss with model size}
\begin{figure}
    \centering
    \includegraphics[scale=0.5]{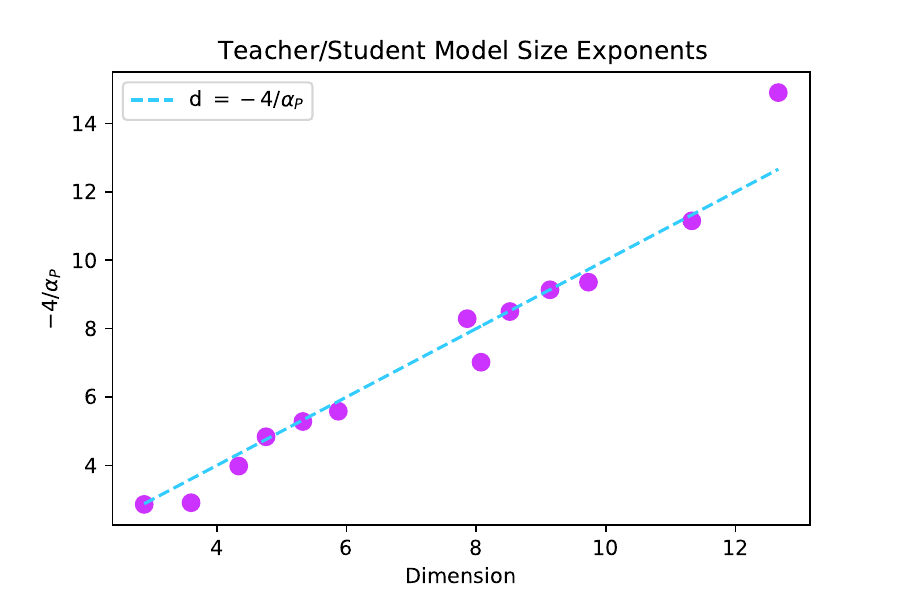}
    \caption{\small This figure shows the variation of $\alpha_P$ with the input-space dimension. The exponent $\alpha_P$ is the scaling exponent of loss with model size for teacher-student setup.}
    \label{fig:TS_ModelSize}
\end{figure}

We replicated the teacher-student setup in \cite{sharma2020neural} to demonstrate the scaling of loss with model size. The resulting variation of $-4/\alpha_P$ with input-space dimensionality is shown in figure \ref{fig:TS_ModelSize}. In our implementation we averaged (in log space) over $15$ iterations, with a fixed, randomly generated teacher.

\subsection{Effect of aspect ratio on scaling exponents}
\begin{figure}
     \centering
     \begin{subfigure}
         \centering
         \includegraphics[width=\textwidth]{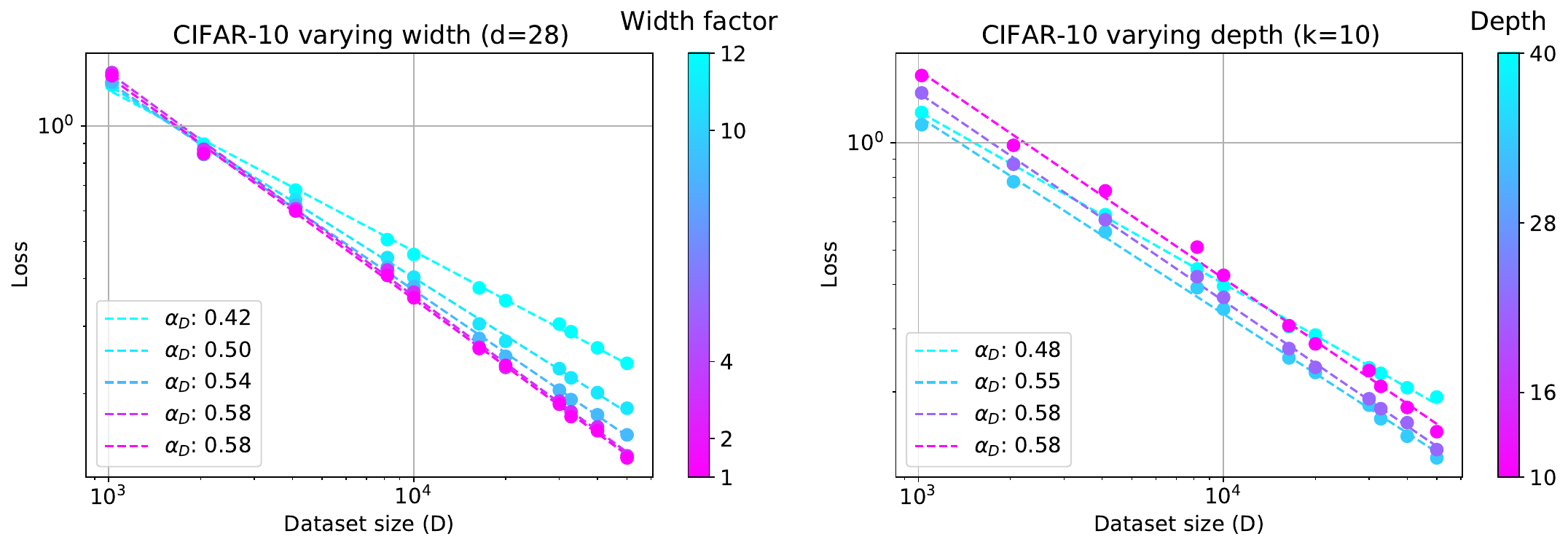}
         %\caption{}
     \end{subfigure}
     \hfill
        \caption{\small\textbf{Effect of aspect ratio on dataset scaling.} We find that for WRN-d-k trained on CIFAR-10, varying depth from 10 to 40 has a relatively mild effect on scaling behavior, while varying the width multiplier, $k$, from 1 to 12 has a more noticeable effect, up until a saturating width.}
        \label{fig:vary_arch}
\end{figure}

We trained Wide ResNet architectures of various widths and depths on CIFAR-10 accross dataset sizes. We found that the effect of depth on dataset scaling was mild for the range studied, while the effect of width impacted the scaling behavior up until a saturating width, after which the scaling behavior fixed. See Figure~\ref{fig:vary_arch}.

\section{Proof of Theorem 1}
We now prove Theorem~\ref{thm:supp_loss_var}, repeated below for convenience.
\begin{supp_thm}\label{thm:supp_loss_var}
Let $\ell(f)$ be the test loss as a function of network output, ($L=\mathbb{E}\left[\ell(f)\right]$),
and let $f_{T}$ be the network output after $T$ training steps,
thought of as a random variable over weight initialization,
draws of the training dataset, and optimization seed. Further
let $f_{T}$ be concentrating with
$\mathbb{E}[\left(f_{T}-\mathbb{E}[f_{T}]\right)^{k}]=\mathcal{O}\left(\epsilon\right) \forall k\geq2$. If $\ell$ is a finite degree polynomial, or has bounded second derivative, or is 2-H\"older, then $\mathbb{E}\left[\ell(f_{T})\right]-\ell\left(\mathbb{E}\left[f_T\right]\right)=\mathcal{O}(\epsilon)$.
\end{supp_thm}
\begin{proof}
\emph{Case 1 -- finite degree polynomial}: In this case, we can write,
\es{eq:supp_polynomial_loss}{
\ell(f_{T})-\ell(\mathbb{E}\left[f_{T}\right])=\sum_{k=1}^{K}\frac{\ell^{(k)}\left(\mathbb{E}\left[f_{T}\right]\right)}{k!}\left(f_{T}-\mathbb{E}\left[f_{T}\right]\right)^{k}\,,
}
where $K$ is the polynomial degree and $\ell^{(k)}$ is the $k$-th derivative of $\ell$. Taking the expectation of \eqref{eq:supp_polynomial_loss} and using the moment scaling proves the result.\\

\noindent \emph{Case 2 -- bounded second derivative}: The quadratic mean value theorem states that for any $f_{T}$, there exists a $c$ such that,
\es{eq:supp_quad_mvt}{
\ell(f_{T})-\ell(\mathbb{E}\left[f_{T}\right]) = \left(f_{T}-\mathbb{E}\left[f_{T}\right]\right)\ell'(\mathbb{E}\left[f_{T}\right])+\frac{1}{2}\ell''(c)\left(f_{T} - \mathbb{E}\left[f_{T}\right]\right)^{2}\,.
}
Taking the expectation of \eqref{eq:supp_quad_mvt} and using the fact that $f''(c)$ is bounded yields the desired result.\\

\noindent \emph{Case 3 -- 2-H\"older}: Lastly, the loss being 2-H\"older means we may write,
\es{eq:supp_holder}{
\ell(f_{T})-\ell(\mathbb{E}\left[f_{T}\right]) \leq \left |\ell(f_{T})-\ell(\mathbb{E}\left[f_{T}\right])\right| \leq K_{\ell}\left(f_{T} - \mathbb{E}\left[f_{T}\right]\right)^{2}\,.
}
Again, taking the expectation of this inequality completes the proof.
\end{proof}

\subsubsection*{Remarks on loss variance} Theorem
%~\ref{thm:loss_var} 
\ref{thm:supp_loss_var} concerns the mean loss, however we would also like to understand if this scaling holds for typical instances. This can be understood by examining how  the variance of the loss or alternatively how $\mathbb{E}\left[\left|\ell\left(f_T\right)-\ell\left(\mathbb{E}\left[f_{T}\right]\right)\right|\right]$ scales.

\noindent For \emph{Case 3 -- 2-H\"older loss}, we can rerun the argument of Theorem~\ref{thm:supp_loss_var}, using \eqref{eq:supp_holder} to yield $\mathbb{E}\left[\left|\ell\left(f_T\right)-\ell\left(\mathbb{E}\left[f_{T}\right]\right)\right|\right]=\mathcal{O}\left(\epsilon\right)$.

\noindent For \emph{Cases 1 and 2}, we can attempt to apply the same argument as in the proof. This \emph{almost} works. In particular, using H\"older's inequality, $\mathbb{E}[\left(f_{T}-\mathbb{E}[f_{T}]\right)^{k}]=\mathcal{O}\left(\epsilon\right) \forall k\geq2$ implies
$\mathbb{E}[\left|f_{T}-\mathbb{E}[f_{T}]\right|^{k}]=\mathcal{O}\left(\epsilon\right) \forall k\geq2$. Taking the absolute value and expectation of \eqref{eq:supp_polynomial_loss} or \eqref{eq:supp_quad_mvt} then gives
\es{eq:supp_almost}{
\mathbb{E}\left[\left|\ell\left(f_T\right)-\ell\left(\mathbb{E}\left[f_{T}\right]\right)\right|\right]\leq \left|\ell'\left(\mathbb{E}\left[f_{T}\right]\right)\right|\mathbb{E}\left[\left|f_{T}-\mathbb{E}\left[f_{T}\right]\right|\right]+\mathcal{O}\left(\epsilon\right)\,.
}

In general, the above assumptions on $\ell$ and $f_{T}$ imply only that $\mathbb{E}\left[\left|f_{T}-\mathbb{E}\left[f_{T}\right]\right|\right]=\mathcal{O}\left(\sqrt{\epsilon}\right)$ and thus typical instances of the loss will exhibit a less dramatic scaling with $\epsilon$ than the mean. If we further assume, however, that $f_{T}$ on average has been trained such as to be sufficiently close to a local minimum of the loss, such that $\left|\ell'\left(\mathbb{E}\left[f_{T}\right]\right)\right|=\mathcal{O}\left(\sqrt{\epsilon}\right)$, then typical instances will also obey the $\mathcal{O}\left(\epsilon\right)$ scaling.

\section{Variance-limited dataset scaling}
In this section, we expand on our discussion of the variance-limited dataset scaling, $L(D)-\lim_{D\rightarrow\infty}L(D)=\mathcal{O}\left(D^{-1}\right)$. We first explain some intuition for why this behavior might be expected for sufficiently smooth loss. We then derive it explicitly for losses that are polynomial in the weights. Finally, we present non-smooth examples where the scaling can be violated either by having unbounded loss, or first derivative.

\subsection{Intuition}
At a high level, the intuition is as follows. For any fixed value of weights, $\theta$, the training loss with $D$ training points (thought of as a random variable over draws of the dataset), $L_{\textrm{train}}[\theta]$ concentrates around the population loss $L_{\textrm{pop}}[\theta]$, with variance that scales as $\mathcal{O}\left(D^{-1}\right)$.

Our optimization procedure can be thought of as a map from initial weights and training loss to final weights $\textrm{Op}: (\theta_{0}, L_{\textrm{train}}[\theta])\rightarrow\theta_{T}$. If this map is sufficiently smooth -- for instance satisfying the assumptions of Theorem 1
%~\ref{thm:loss_var}
or well approximated by a Taylor series about all $\mathbb{E}_{D}\left[L_{\textrm{train}}[\theta_t]\right]$ -- then the output, $\theta_{T}$, will also concentrate around its infinite $D$ limit with variance scaling as $\mathcal{O}\left(D^{-1}\right)$. Finally, if the population loss is also sufficiently smooth, the test loss for a model trained on $D$ data points averaged over draws of the dataset, $L(D)=\mathbb{E}_{D}\left[L_{\textrm{pop}}[\theta_{T}]\right]$, satisfies $L(D)-\lim_{D\rightarrow\infty}L(D)=\mathcal{O}\left(D^{-1}\right)$. We now walk through this in a little more detail.

\subsubsection*{Early time}
We can follow this intuition a bit more explicitly for the first few steps of gradient descent. As the training loss at initialization, $L_{\textrm{train}}[\theta_{0}]$, is a sample average over $D$ i.i.d draws, it concentrates around the population loss $L_{\textrm{pop}}[\theta_{0}]$ with variance $\mathcal{O}\left(D^{-1}\right)$. As a result, the initial gradient, $g_{0}=\frac{\partial L_{\textrm{train}}}{\partial\theta_0}$ will also concentrate with $O\left(D^{-1}\right)$ variance and so will the weights at time step $1$, $\theta_{1}=\theta_0-\eta g_{0}$. The training loss at time step 1 is then given by (setting $\eta = 1$),
\es{eq:supp_L_train_t1}{
L_{\textrm{train}}[\theta_1]&=L_{\textrm{train}}[\theta_0- g_{0}]\,.
}
If $L_{\textrm{train}}$ is sufficiently smooth around $\theta_0-\mathbb{E}_{D}\left[g_{0}\right]$, then we get that $L_{\textrm{train}}[\theta_1]$ concentrates around $L_{\textrm{train}}[\theta_1]$ with $O\left(D^{-1}\right)$ variance. We can keep bouncing back and forth between gradient (or equivalently weights) and training loss for any number of steps $T$ which does not scale with $D$. Plugging this final $\theta_{T}$ into the population loss and taking the expectation over draws of the training set, $L(D)=\mathbb{E}_{D}\left[L_{\textrm{pop}}[\theta_{T}]\right]$. If $L_{\textrm{pop}}$ is also sufficiently smooth, this yields $L(D)-\lim_{D\rightarrow\infty}L(D)=\mathcal{O}\left(D^{-1}\right)$. 

Here we have used the term sufficiently smooth. A sufficient set of criteria are given in Theorem 1;
%~\ref{thm:loss_var}
however, this is likely too restrictive. Indeed, any set of train and population loss for which a Taylor series (or asymptomatic series with optimal truncation) give an $O\left(D^{-1}\right)$ error around the training points $\mathbb{E}_{D}\left[\theta_{t=0\ldots T}\right]$ will have this behavior.

\subsubsection*{Local minimum}
The above intuition relied on training for a number of steps that was fixed as $D$ is taken large. Here we present some alternative intuition for the variance-limited scaling at late times, as training approaches a local minimum in the loss. For simplicity we discuss a one-dimensional loss.

Consider a local minimum, $\theta^{*}$, of the population loss. As $D$ is taken large, with high probability, the training loss will have a local minimum, $\bar{\theta}^{*}$, such that $|\theta^{*}-\bar{\theta}^{*}|=\mathcal{O}\left(D^{-1}\right)$. One way to see this is to note that for a generic local minimum the first derivative changes sign, i.e. we can find $\theta_{1}, \theta_{2}$ such that $\theta_{1}<\theta^{*}<\theta_{2}$ and either $L'_{\textrm{pop}}[\theta_{1}]<0, L'_{\textrm{pop}}[\theta_{2}]>0$ or $L'_{\textrm{pop}}[\theta_{2}]<0, L'_{\textrm{pop}}[\theta_{1}]>0$. To be concrete let's focus on the first case (the argument will be identical in either case). As $D$ becomes large, the probability that the training loss at $\theta_1$ and $\theta_{2}$ differs significantly from the population loss approaches zero. This can be seen from Markov's inequality, where, 
$P\left(\left|L'_{\textrm{train}}[\theta]-L'_{\textrm{pop}}[\theta]\right|>a\right)\leq\frac{\textrm{Var}_{D}\left(L'_{\textrm{train}}[\theta]\right)}{a^{2}}
$,
or more dramatically from Hoeffding's inequality (assuming bounded $L_{\textrm{train}}-L_{\textrm{pop}}$ lying in an interval of size $I$)
\es{eq:subb_hoeffding}{
P\left(\left|L'_{\textrm{train}}[\theta]-L'_{\textrm{pop}}[\theta]\right|>a\right)\leq2e^{-\frac{2}{I} D^{2}a^{2}}\,.
}
Here to have non-vanishing probability as we take $D$ large, $L'_{\textrm{train}}[\theta_{1}]$ and $L'_{\textrm{train}}[\theta_{2}]$ must be closer than $\mathcal{O}\left(D^{-1}\right)$. If $\theta_{1}$ and $\theta_2$ are taken to be $\mathcal{O}\left(D^{0}\right)$, then $L'_{\textrm{train}}$ must change sign, indicating an extremum of $L_{\textrm{train}}$; however, we can do even better. If we assume $L_{\textrm{train}}$ is Lipshitz about $\theta^*$ then we can still ensure a sign change even if $|\theta_1-\theta^{*}|,|\theta_2-\theta^*|=\mathcal{O}\left(D^{-1}\right)$. Using concentration of $L''_{\textrm{train}}[\theta]$ ensures the extremum is a local minimum. For non-generic minimum (i.e. vanishing first derivatives) we can apply the same arguments to higher order derivatives (assuming they exist) of $L_{\textrm{pop}}.$ Thus for a local minimum of $L_{\textrm{pop}}$, with high probability $L_{\textrm{train}}$ will have a corresponding minimum within a distance $\mathcal{O}\left(D^{-1}\right)$.

If we now consider an initialization $\theta_0$ and training procedure such that training converges to the local minimum of the training loss, $\bar{\theta}^{*}$, and that the population loss is sufficiently smooth about $\theta^{*}$ (e.g. Lipshitz), then $\mathbb{E}_{D}[L_{\textrm{train}}[\bar{\theta}^{*}]-L_{\textrm{pop}}[\theta^{*}]]=\mathbb{E}_{D}[L_{\textrm{train}}[\bar{\theta}^{*}]-L_{\textrm{pop}}[\bar{\theta}^{*}]]+\mathbb{E}_{D}[L_{\textrm{pop}}[\bar{\theta}^{*}]-L_{\textrm{pop}}[\theta^{*}]]$. The first term vanishes, while the second is $\mathcal{O}(D^{-1})$. If we further assume that this happens on average over choices of $\theta_0$ then we expect $L(D)-\lim_{D\rightarrow\infty}L(D)=\mathcal{O}\left(D^{-1}\right)$.

\subsubsection*{Stochastic gradient descent (SGD)}
At first blush it may be surprising that the variance-limited scaling holds even for mini-batch training. Indeed in this case, there is batch noise that comes in at a much higher scale than any variance due to the finite training set size. 
Indeed, the effect of mini-batching changes the final test loss, however if we fix the SGD procedure or average over SGD seeds, as we take $D$ large, we can still ask how the training loss for a model trained under SGD on a training set of size $D$ differs from that for a model trained under SGD on an infinite training set.

To see this, we first consider averaging over minibatches of size $B$, but where points are drawn i.i.d. with replacement. If we denote the batch at step $t$ by $\mathcal{B}_t$ and the average over independent draws of this batch by $\mathbb{E}_{B}\left[\bullet\right]$, then note we can translate moments with respect to batch draws with empirical averages over the entire training set. Explicitly, consider $c_a$ and $d_a$ potentially correlated, but each drawn i.i.d. within a batch. We have that,
\es{eq:supp_batch_to_emp}{
\mathbb{E}_B\left[\frac{1}{B}\sum_{a\in\mathcal{B}_t}c_{a}\right]&=\frac{1}{D}\sum_{a=1}^{D}c_{a}\\
\mathbb{E}_B\left[\left(\frac{1}{B}\sum_{a\in\mathcal{B}_t}c_{a}\right)\left(\frac{1}{B}\sum_{a'\in\mathcal{B}_t}d_{a'}\right)\right]&=\left(1-\frac{1}{B}\right)\left(\frac{1}{D}\sum_{a=1}^{D}c_{a}\right)\left(\frac{1}{D}\sum_{a'=1}^{D}d_{a'}\right)+\frac{1}{B}\frac{1}{D}\sum_{a=1}^{D}c_{a}d_a\,.
}
This procedure means, after taking an average over draws of SGD batch, rather than thinking about a function of mini-batch averages, we can equivalently consider a modified function, with explicit dependence on the batch size, but that is only a function of empirical means over the training set. We can thus recycle the above intuition for the scaling of smooth functions of empirical means. 

The above relied on independently drawing every sample from every batch. At the other extreme, we can consider drawing batches without shuffling and increasing training set size by $B$ datapoints at a time, so as to keep the initial set of batches in an epoch fixed. In this case, the first deviation in training between a dataset of size $D$ and one of size $D+B$ happens at the last batch in the first epoch after processing $D$ datapoints. 

As an extreme example, consider the case where $D>BT$. In this case, as we only take $T$ steps, the loss is constant for all $D>BT$ and so $\lim_{D\rightarrow\infty}L(D;T;B)=L(BT;T;B)$ and thus $L(D>BT)-\lim_{D\rightarrow\infty}L(D)=0$ (and in particular is trivially $\mathcal{O}\left(D^{-1}\right)$).

\subsection{Polynomial loss}
Before discussing neural network training, we review the concentration behavior of polynomials of sample means.
\begin{lemma}\label{lemma:supp_moments}
Let $\bar{c}^{(i)}=\frac{1}{D}\sum_{a=1}^{D}c^{(i)}_{a}$ for $i=0\ldots J$ be empirical means, over $D$ i.i.d. draws of $c^{(i)}_{a}$ and let $c^{(i)}$ denote the distributional mean. Further, let
$X=(\bar{c}^{(0)})^{k_{0}}(\bar{c}^{(1)})^{k_{1}}\cdots(\bar{c}^{(J)})^{k_{J}}$ be a monomial in the sample means. Then $X$ concentrates with moments $\mathcal{O}\left(D^{-1}\right)$,
\es{eq:supp_lemma_moments}{
\mathbb{E}_{D}\left[\left(X-({c}^{(0)})^{k_{0}}({c}^{(1)})^{k_{1}}\cdots({c}^{(J)})^{k_{J}}\right)^{n}\right]=\mathcal{O}\left(D^{-1}\right)\,.}
Here, $\mathbb{E}_{D}\left[\bullet\right]$ denotes the average over independent draws of $D$ samples.
\end{lemma}
\begin{proof} To establish this we can proceed by direct computation.
\es{eq:supp_binomial}{
&\mathbb{E}_{D}\left[\left(X-({c}^{(0)})^{k_{0}}({c}^{(1)})^{k_{1}}\cdots({c}^{(J)})^{k_{J}}\right)^{n}\right]\\
&\qquad=\sum_{p=0}^{n}(-1)^{n-p}
\left(
\begin{array}{c}
     n\\
     p 
\end{array}
\right)\mathbb{E}_{D}\left[X^{p}\right]\left(({c}^{(0)})^{k_{0}}({c}^{(1)})^{k_{1}}\cdots({c}^{(J)})^{k_{J}}\right)^{n-p}
}
Each term in the sum can be computed using
\es{eq:supp_sum_counting}{
\mathbb{E}_{D}\left[X^{p}\right]&=
\mathbb{E}_{D}\left[
(\bar{c}^{(0)})^{pk_{0}}(\bar{c}^{(1)})^{pk_{1}}\cdots(\bar{c}^{(J)})^{pk_{J}}
\right]\\
&=
\frac{1}{D^{(p\sum_{i=0}^{J}k_{i})}}\sum_{\{a^{(i)}_{\alpha}\}}
\mathbb{E}_{D}\left[
\left(c_{a^{(0)}_{1}}^{(0)}\cdots c_{a^{(0)}_{pk_{0}}}^{(0)}\right)\left(c_{a^{(1)}_{1}}^{(1)}\cdots c_{a^{(1)}_{pk_{1}}}^{(1)}\right)\cdots\left(c_{a^{(J)}_{1}}^{(J)}\cdots c_{a^{(J)}_{pk_{J}}}^{(J)}\right)
\right]\\
&=
\frac{1}{D^{(p\sum_{i=0}^{J}k_{i})}}\sum_{\{a^{(i)}_{\alpha}\neq a^{(j)}_{\beta}\}}
\mathbb{E}_{D}\left[
\left(c_{a^{(0)}_{1}}^{(0)}\cdots c_{a^{(0)}_{pk_{0}}}^{(0)}\right)\left(c_{a^{(1)}_{1}}^{(1)}\cdots c_{a^{(1)}_{pk_{1}}}^{(1)}\right)\cdots\left(c_{a^{(J)}_{1}}^{(J)}\cdots c_{a^{(J)}_{pk_{J}}}^{(J)}\right)
\right]\\
&\qquad\qquad+\mathcal{O}\left(D^{-1}\right)\\
&=
\frac{D(D-1)\cdots(D-(p\sum_{i=0}^{J}k_{i}-1))}{D^{(p\sum_{i=0}^{J}k_{i})}}\left(c^{(0)}\right)^{p k_{0}}\left(c^{(1)}\right)^{p k_{1}}\cdots\left(c^{(J)}\right)^{p k_{J}}+\mathcal{O}\left(D^{-1}\right)\\
&=\left(\left(c^{(0)}\right)^{k_{0}}\left(c^{(1)}\right)^{k_{1}}\cdots\left(c^{(J)}\right)^{k_{J}}\right)^{p}+\mathcal{O}\left(D^{-1}\right)\,.\nonumber
}
Plugging this into \eqref{eq:supp_binomial} establishes the lemma. 
\end{proof}
In the above, we use the multi-index notation $\{a^{(i)}_{\alpha}\}$ for the collection of indices on the $c^{i}$ and the notation $\{a^{(i)}_{\alpha}\neq a^{(j)}_{\beta}\}$ for the subset of terms in the sum where all indices take different values. 

Lemma~\ref{lemma:supp_moments} immediately implies that the mean of polynomials of $\bar{c}^{(i)}$ concentrate around their infinite data limit.
\es{}{
\mathbb{E}_{D}\left[\left(g\left(\bar{c}^{(0)},\bar{c}^{(1)},\ldots,\bar{c}^{(K)}\right)-g\left({c}^{(0)},{c}^{(1)},\ldots,{c}^{(K)}\right)\right)^{n}\right]=\mathcal{O}\left(D^{-1}\right)\,,   
}
for $g\in P_{K}\left[\bar{c}^{(0)},\bar{c}^{(1)},\ldots,\bar{c}^{(K)}\right]$.

With this out of the way, we can proceed to analyzing the scaling of trained neural networks. Here we consider the simplified setting where the network map, $f$, and loss $\ell$ evaluated on each training example, $\mathbf{x}_{a}=(x_{a},y_{a})$, are polynomial of degree $J$ and $K$ in the weights, $\theta_{\mu}$,
\es{poly_sample_loss}{
f(x)&=\sum_{i=1}^{J}b^{(i)}_{\mu_{1}\mu_{2}\ldots\mu_{i}}(x)\theta_{\mu_{1}}\theta_{\mu_{2}}\cdots\theta_{\mu_{i}}\, \ \ \ \ell(\mathbf{x}_{a})\,=\,\sum_{i=1}^{K}c^{(i)}_{\mu_{1}\mu_{2}\ldots\mu_{i}}(\mathbf{x}_{a})\theta_{\mu_{1}}\theta_{\mu_{2}}\cdots\theta_{\mu_{i}}\,.
}
The training loss can then be written as,
\es{poly_training_loss}{
L_\textrm{train}&=\sum_{i=1}^{K}\bar{c}^{(i)}_{\mu_{1}\mu_{2}\ldots\mu_{i}}\theta_{\mu_{1}}\theta_{\mu_{2}}\cdots\theta_{\mu_{i}}\,, \ \ \ \bar{c}^{(i)}\,=\,\frac{1}{D}\sum_{a=1}^{D}c^{(i)}(\mathbf{x}_{a})\,.
}
Here we have used the convention that the repeated weight indices $\mu_{j}$ are summed over.

\subsubsection*{Gradient descent} As a result of the gradient descent weight update, $\theta_{t+1}=\theta_{t}-\eta\frac{\partial L_{\textrm{train}}}{\partial\theta}$, the weights at time $T$ are a polynomial of degree $\left(K-1\right)^{T}$ in the $\bar{c}^{(i)}$.
\es{poly_gd_weights}{
\theta_{T}\in P_{\left(K-1\right)^{T}}\left[\bar{c}^{(0)},\bar{c}^{(1)},\ldots,\bar{c}^{(K)}\right]\,.} The coefficients of this polynomial depend on the initial weights, $\theta_{0}$. Plugging these weights back into the network output, we have that the network function at time $T$ is again a polynomial in $\bar{c}^{(i)}$, now with degree $J\left(K-1\right)^{T}$.
\es{poly_gd_map}{
f_{T}(x)\in P_{J\left(K-1\right)^{T}}\left[\bar{c}^{(0)},\bar{c}^{(1)},\ldots,\bar{c}^{(K)}\right]\,.}
Thus, again using Lemma~\ref{lemma:supp_moments}, $f_{T}$ concentrates with variance $\mathcal{O}(D^{-1})$. 
\es{eq:supp_f_var}{
\mathbb{E}_{D}\left[\left(f_{T}-\mathbb{E}_{D}\left[f_{T}\right]\right)^{2}\right]=\mathcal{O}\left(D^{-1}\right)\,.
}
and by Theorem 1 the loss will obey they variance-limited scaling.

\subsubsection*{Stochastic gradient descent} We now consider the same setup of polynomial loss, but now trained via stochastic gradient descent (SGD). We consider SGD batches drawn i.i.d. with replacement and are interested in the test loss averaged over SGD draws, with fixed batch size, $B$.

We proceed by proving the following lemma, which allows us to reuse a similar argument to the GD case.
\begin{lemma}\label{lemma:supp_sgd_moments}
Let $\tilde{c}^{(i;t)}=\frac{1}{B}\sum_{a\in\mathcal{B}_{t}}c^{(i)}_{a}$ for $i=0\ldots J$ be mini-batch averages, over $B$ i.i.d. draws of $c^{(i)}_{a}$.
% and let $\bar{c}^{(i)}$ denote the empirical mean over the full training set as in Lemma~\ref{lemma:supp_moments}.
Further, let $X=(\tilde{c}^{(0;t_{0})})^{k_{0}}(\tilde{c}^{(1;t_{1})})^{k_{1}}\cdots(\tilde{c}^{(J;t_{J})})^{k_{J}}$ be a monomial in the mini-batch means. Then $\mathbb{E}_{B}\left[X\right]\in P_{\sum_{i=0}^{J}k_{i}}\left[\bar{d}^{(0)},\bar{d}^{(1)},\ldots,\bar{d}^{(\prod_{i=0}^{J}(k_{i}+1)-1)}\right]$, where $\bar{d}^{(i)}$ are empirical means over the full training set of i.i.d. random variables as in Lemma~\ref{lemma:supp_moments} and $\mathbb{E}_{B}\left[\bullet\right]$ denotes the expectation over draws of SGD batches of size $B$.
\end{lemma}
\begin{proof} Expectations over draws of batches at different time steps are independent. Thus, w.l.o.g., we can consider $t:=t_{0}=t_{1}=\cdots=t_{J}$. We can again proceed by direct computation, expanding the mini-batch sums,
\es{eq:supp_expand_sgd}{
\mathbb{E}_{B}\left[X\right]&=\frac{1}{B^{\sum_{i=0}^{J}k_{i}}}\mathbb{E}_{B}\left[\sum_{\{a_{\alpha}^{(i)}\}\in\mathcal{B}_{t}}\left(c_{a_{1}^{(0)}}^{(0)}\cdots c_{a_{k_{0}}^{(0)}}^{(0)}\right)\left(c_{a_{1}^{(1)}}^{(1)}\cdots c_{a_{k_{1}}^{(1)}}^{(1)}\right)\cdots \left(c_{a_{1}^{(J)}}^{(J)}\cdots c_{a_{k_{J}}^{(J)}}^{(J)}\right)\right]\,.
}
To proceed, we must keep track of terms in the sum where the $a^{(i)}_{\alpha}$ take the same or different values. If all $a^{(i)}_{\alpha}$ are different, the expectation over batch draws fully factorizes. More generally \eqref{eq:supp_expand_sgd} can be decomposed as a sum over products. 

One way of keeping track of the index combinatorics is to introduce a set of graphs, $\Gamma$, where each graph $\gamma\in\Gamma$ has $k_{0}$ vertices of type 0, $k_{1}$ vertices of type 1, \ldots, and $k_{J}$ vertices of type J (one vertex for each $a^{(i)}_{\alpha}$ index). Any pair of vertices may have zero or one edge between them. For any set of three vertices, $v_{1}$, $v_{2}$, and $v_{3}$ with edges $(v_{1},v_{2})$ and $(v_{2},v_{3})$ there must also be an edge $(v_{1},v_{3})$. The set $\Gamma$ consists of all possible ways of connecting these vertices consistent with these rules.

For each graph, $\gamma$, we denote connected components by $\sigma$ and denote the number of vertices of type $i$ within the connected component $\sigma$ by $m_{\sigma}^{(i)}$. With this we can write the sum, \eqref{eq:supp_expand_sgd} as
\es{eq:supp_expand_sgd_graph}{
\mathbb{E}_{B}\left[X\right]&=\sum_{\gamma\in\Gamma}S_{\gamma}(B)\prod_{\sigma\in\gamma}\mathbb{E}_{B}\left[\frac{1}{B}\sum_{a\in\mathcal{B}_{t}}\left(c^{(0)}_a\right)^{m_{\sigma}^{(0)}}\left(c^{(1)}_a\right)^{m_{\sigma}^{(1)}}\cdots\left(c^{(J)}_a\right)^{m_{\sigma}^{(J)}}\right]\\
&=\sum_{\gamma\in\Gamma}S_{\gamma}(B)\prod_{\sigma\in\gamma}\frac{1}{D}\sum_{a=1}^{D}\left(c^{(0)}_a\right)^{m_{\sigma}^{(0)}}\left(c^{(1)}_a\right)^{m_{\sigma}^{(1)}}\cdots\left(c^{(J)}_a\right)^{m_{\sigma}^{(J)}}\\
&=\sum_{\gamma\in\Gamma}S_{\gamma}(B)\prod_{\sigma\in\gamma}\bar{d}^{(\{m_{\sigma}^{(0)},m_{\sigma}^{(1)},\ldots,m_{\sigma}^{(J)}\})}\,.}
Here $S_{\gamma}(B)$ is a combinatoric factor associated to each graph, not relevant for the argument. The $m_{\sigma}^{(i)}$ take on values $0$ to $k_i$, so the multi-index can take on $\prod_{i=1}^{J}(k_{i}+1)$ different values, which we re-index to $\bar{d}^{(0)},\bar{d}^{(1)},\ldots,\bar{d}^{(\prod_{i=1}^{J}(k_{i}+1)-1)}$. Meanwhile, the degree of \eqref{eq:supp_expand_sgd_graph} in $\bar{d}^{(i)}$ is bounded by the number of total vertices in each graph, i.e. $\sum_{i=0}^{J}k_i$. This establishes the lemma.
\end{proof}
For a polynomial loss of degree $K$, the mini-batch training loss at each time step takes the form
\es{eq:supp_sgd_loss}{
L_\textrm{train}^{(t)}&=\sum_{i=1}^{K}\tilde{c}^{(i;t)}_{\mu_{1}\mu_{2}\ldots\mu_{i}}\theta_{\mu_{1}}\theta_{\mu_{2}}\cdots\theta_{\mu_{i}}\,, \ \ \ \tilde{c}^{(i;t)}\,=\,\frac{1}{B}\sum_{a=\in\mathcal{B}_{t}}c^{(i)}(\mathbf{x}_{a})\,.
}
The update rule, $\theta_{t+1}=\theta_{t}-\eta\frac{\partial L_{\textrm{train}}^{(t+1)}}{\partial\theta}$ ensures that $\theta_{T}$ is a polynomial of degree $(K-1)^{T}$ in the $\tilde{c}^{(i;0)},\tilde{c}^{(i;1)},\cdots,\tilde{c}^{(i;T)}$
\es{eq:supp_theta_poly_sgd}{
\theta_{T}\in P_{(K-1)^{T}}\left[\tilde{c}^{(0;0)},\tilde{c}^{(0;1)},\ldots,\tilde{c}^{(0;T)},\tilde{c}^{(1;0)},\tilde{c}^{(1;1)},\ldots,\tilde{c}^{(1;T)},\ldots,\tilde{c}^{(K;0)},\tilde{c}^{(K;1)},\ldots,\tilde{c}^{(K;T)}\right]\,,
}
and consequently, denoting the test loss evaluated at $\theta_{T}$ by $L[\theta_{T}]$,
\es{eq:supp_L_poly_sgd}{
L[\theta_{T}]\in P_{K(K-1)^{T}}\left[\tilde{c}^{(0;0)},\tilde{c}^{(0;1)},\ldots,\tilde{c}^{(0;T)},\tilde{c}^{(1;0)},\tilde{c}^{(1;1)},\ldots,\tilde{c}^{(1;T)},\ldots,\tilde{c}^{(K;0)},\tilde{c}^{(K;1)},\ldots,\tilde{c}^{(K;T)}\right]\,.
}
Using Lemma~\ref{lemma:supp_sgd_moments}, the expectation of $L[\theta_{T}]$ over draws of SGD batches is given by
\es{eq:supp_L_poly_sgd_mean}{
\mathbb{E}_{B}\left[L[\theta_{T}]\right]\in P_{K(K-1)^{T}}\left[\bar{d}^{(0)},\ldots,\bar{d}^{(K^{K}(K-1)^{TK})}\right]\,.
}
Finally, denoting $\mathbb{E}_{D}\left[\mathbb{E}_{B}\left[L[\theta_{T}]\right]\right]$ by $L(D;B)$ and applying Lemma~\ref{lemma:supp_moments} gives
\es{eq:subb_loss_varlimit_sgd}{
L(D;B)-\lim_{D\rightarrow\infty}L(D;B)=\mathcal{O}\left(D^{-1}\right)\,.
}
\subsection{Non-smooth examples}
Here we present two worked examples where non-bounded or non-smooth loss leads to violations of the variance dominated scaling. In example one, the system obeys the variance dominated scaling at early times, but exhibits different behavior for times larger than the dataset size. In the second example, the system violates the variance dominated scaling even for two gradient descent steps, as a result of an unbounded derivative in the loss.

\subsubsection*{Example 1 -- unbounded loss at late times}
Consider a dataset with two varieties of data points, drawn with probabilities $\alpha$ and $1-\alpha$, and one-dimensional quadratic losses, $\ell_1$ (concave up) and $\ell_{2}$ (concave down), on these two varieties.
\es{eq:supp_ex1_point_loss}{
\ell_{1}(\theta)\,=\,\frac{1}{2}\theta^{2}\,, \ \ \ \ell_{2}(\theta)\,=\,-\frac{1}{2}\theta^{2}\,.
}
If, in a slight abuse of notation, we further denote the training loss on a sample with $n_{1}$ points of type 1 and $D-n_1$ points of type two by $\ell_{n_{1}}$ and the population loss at a given value of the weight by $L_{\textrm{pop}}$, we have
\es{eq:supp_ex1_train_loss}{
\ell_{n_1}\,=\,\left(\frac{n_{1}}{D}-\frac{1}{2}\right)\theta^{2}\,, \ \ \ L_{\textrm{pop}}\,=\,\left(\alpha-\frac{1}{2}\right)\theta^{2}\,.
}
For this example we take $\alpha>1/2$. In this case, the minimum of the population loss is at zero,  while the minimum of the training loss can be at zero or at $\pm\infty$ depending on whether the training sample has $n_1$ greater than or less than $D/2$. We can thus create a situation where at late training times, $\theta_{T}$ does not concentrate around the minimum of the population loss.

As we work through this example explicitly, we will see the following. (i) A mismatch larger than $\mathcal{O}\left(D^{-1}\right)$ between the population minimum and the minimum found by training on a sample set of size $D$ requires times $T$ larger than a constant multiple of $D$. (ii) The quantity we study throughout this work is the difference between the infinite data limit of the test loss and the finite data value, $L(D)-\lim_{D\rightarrow\infty}L(D)$. The minimum of the infinite data limit of the test loss is not the same as the minimum of the population loss, $\min \lim_{D\rightarrow\infty}L(D)\neq \min_{\theta}L_{\textrm{pop}}$. In this example one diverges, while the other is finite. In particular this example evades the scaling result by $L(D)$ for times larger than $D$ having a diverging limit. 

Explicitly, we study the evolution of the model under gradient flow.
\es{eq:supp_ex1_gflow}{
\dot{\theta}\,=\,-2\left(\frac{n_{1}}{D}-\frac{1}{2}\right)\theta\,, \ \ \ \theta_{T}\,=\,e^{-2\left(\frac{n_1}{D}-\frac{1}{2}\right)T}\theta_{0}\,.
}
The test loss averaged over draws of the dataset is given by
\es{eq:supp_ex1_ltest}{
L(D;T)&=\mathbb{E}_{n_1}\left[\left(\alpha-\frac{1}{2}\right)\theta_{T}^{2}\right]\,=\,e^{2T}\left(\alpha-\frac{1}{2}\right)\left(1-\alpha\left(1-e^{-\frac{4T}{D}}\right)\right)^{D}\theta_{0}^{2}
}
If we consider this loss at large $D$ and fixed $T$ we get
\es{eq:supp_ex1_ltest_fixedT}{
L(D;T)&=e^{-4\left(\alpha-\frac{1}{2}\right)T}\left(\alpha-\frac{1}{2}\right)\theta_{0}^{2}\left(1+\frac{8T^{2}\alpha(1-\alpha)}{D}+\mathcal{O}\left(D^{-2}\right)\right)\,,
}
and thus $L(D;T)-\lim_{D\rightarrow\infty}L(D;T)=\mathcal{O}\left(D^{-1}\right)$ as expected. 

If on the other hand we consider taking $T \gg D$ we have
\es{eq:supp_ex1_ltest_Tlarge}{
L(D;T\gg D)&=e^{2T}\left(\alpha-\frac{1}{2}\right)\left(1-\alpha\right)^{D}\theta_{0}^{2}\,,
}
the limit $\lim_{D,T\rightarrow\infty}L(D;T\gg D)$ diverges.

Lastly, we note that if we take $T=\beta D$ with $\beta < |\log(1-\alpha)|/2$ we can approach the large $D$ limit with non-generic, tuneable exponential convergence.  

\subsubsection*{Example 2 -- unbounded derivative}
Again, consider a two variety setup, this case with equal probabilities and per sample losses,
\es{eq:supp_ex2_point_loss}{
\ell_{1}(\theta)\,=\,\frac{1}{2}\theta^{2}+\frac{1}{2\alpha}|\theta|^{\alpha}\,, \ \ \ \ell_{2}(\theta)\,=\,\frac{1}{2}\theta^{2}-\frac{1}{2\alpha}|\theta|^{\alpha}\,.
}
We will consider different values of $\alpha>0$. The train loss and population loss are then,
\es{eq:supp_ex2_train_loss}{
\ell_{n_1}\,=\,\frac{1}{2}\theta^{2}+\frac{1}{\alpha}\left(\frac{n_{1}}{D}-\frac{1}{2}\right)|\theta|^{\alpha}\,, \ \ \ L_{\textrm{pop}}\,=\,\frac{1}{2}\theta^{2}\,.
}
We consider a model initialized to $\theta_0=1$ and trained for two steps of gradient descent with learning rate 1.
\es{eq:supp_ex2_grad}{
g_{t}\,=\,\theta_{t}+\left(\frac{n_{1}}{D}-\frac{1}{2}\right)\theta_{t}|\theta_{t}|^{\alpha-2}\,, \ \ \ \theta_{t+1}=\theta_{t}-g_{t}\,.
}
Two update steps gives
\es{eq:supp_ex2_theta1}{
\theta_{2}=\left|\frac{n_{1}}{D}-\frac{1}{2}\right|^{\alpha}\,.
}
The test loss is given by the population loss evaluated at $\theta_2$ averaged over test set draws.
\es{eq:supp_ex2_ltest}{
L(D)&=\mathbb{E}_{n_1}\left[\frac{1}{2}\theta_{2}^{2}\right]\,=\,\frac{1}{2^{D+1}}\sum_{n_1=0}^{D}
\left(
\begin{array}{c}
     D \\
     n_1
\end{array}
\right)\left|\frac{n_{1}}{D}-\frac{1}{2}\right|^{2\alpha}\\
&=\sqrt{\frac{D}{2\pi}}\int_{-\infty}^{\infty}e^{-2 D\left(x-\frac{1}{2}\right)^{2}}\left|x-\frac{1}{2}\right|^{2\alpha}+\mathcal{O}\left(D^{-1}\right)\\
&=\frac{\Gamma\left(\alpha+\frac{1}{2}\right)}{2^{1+\alpha}\sqrt{\pi}}D^{-\alpha}+\mathcal{O}\left(D^{-1}\right)\,.\\
}
Here we have approximated the binomial distribution at large $D$ with a normal distribution using Stirling's approximation. 

Note that if $\alpha\geq 1$ then $L(D)-\lim_{D\rightarrow\infty}L(D)=\mathcal{O}\left(D^{-1}\right)$ i.e. the finite sample loss approaches the infinite data loss with the predicted variance-limited scaling. For $0<\alpha<1$, we get a different scaling controlled by $\alpha$. Note that the gradient, expression \eqref{eq:supp_ex2_grad} is singular at the origin for $\alpha$ precisely in this range. 

In summary, this example achieves a different scaling exponent through a diverging gradient. 
%%%%%%%%%%%%%%%%%%%%%%%%%%
\section{Proof of Theorems 2 and 3}\label{sec:supp_proof}
In this section we detail the proof of Theorems 2 and 3.
%~\ref{thm:interp_data} and \ref{thm:interp_params}. 
The key observation is to make use of the fact that nearest neighbor distances for $D$ points sampled i.i.d. from a $d$-dimensional manifold have mean $\lexpp{D,x}\left|x-\hat{x}\right|\rexp=\cO\left(D^{-1/d}\right)$, where $\hat{x}$ is the nearest neighbor of $x$ and the expectation is the mean over datapoints and draws of the dataset, see e.g. \citep{levina2005maximum}.

The theorem statements are copied for convenience. In the main text, in an abuse of notation, we used $L(f)$ to indicate the value of the test loss as a function of the network $f$, and $L(D)$ to indicate the test loss averaged over the population, draws of the dataset, model initializations and training. To be more explicit below, we will use the notation $\ell(f(x))$ to indicate the test loss for a single network evaluated at single test point.

\begin{supp_thm}\label{thm:supp_interp_data}
Let $\ell(f)$, $f$ and $\mathcal{F}$ be Lipschitz with constants $K_{L}$, $K_{f}$, and $K_{\mathcal{F}}$ and $\ell(\mathcal{F})=0$. Further let $\mathcal{D}$ be a training dataset of size $D$ sampled i.i.d from $\mathcal{M}_{d}$ and let $f(x)=\mathcal{F}(x)\,\, \forall x \in\mathcal{D}$,
then $L(D)=\cO\left(K_{L}\textrm{max}{(K_{f},K_{\mathcal{F}})}D^{-1/d}\right)$.
\end{supp_thm}
\begin{proof}
Consider a network trained on a particular draw of the training data. For each training point, $x$, let $\hat{x}$ denote the neighboring training data point. Then by the above Lipschitz assumptions and the vanishing of the loss on the true target, we have $\ell(f(x))\leq K_{L}\left|f(x)-\mathcal{F}(x)\right|\leq K_{L}\left(K_{f}+K_{\mathcal{F}}\right)\left|x-\hat{x}\right|$. With this, the average test loss is bounded as
\es{eq:supp_thm1_bnd}{
L(D)\leq K_{L}\left(K_{f}+K_{\mathcal{F}}\right)\lexpp{D,x}\left|x-\hat{x}\right|\rexp=\cO\left(K_{L}\textrm{max}{(K_{f},K_{\mathcal{F}})}D^{-1/d}\right)\,.
}
In the last equality, we used the above mentioned scaling of nearest neighbor distances.
\end{proof}
\begin{supp_thm}\label{thm:supp_interp_params}
Let $\ell(f)$, $f$ and $\mathcal{F}$ be Lipschitz with constants $K_{L}$, $K_{f}$, and $K_{\mathcal{F}}$. Further let $f(x)=\mathcal{F}(x)$ for $P$ points sampled i.i.d from $\mathcal{M}_{d}$
then $L(P)=\cO\left(K_{L}\textrm{max}{(K_{f},K_{\mathcal{F}})}P^{-1/d}\right)$.
\end{supp_thm}
\begin{proof}
Denote by $\mathcal{P}$ the $P$ points, $z$, for which $f(z)=\mathcal{F}(z)$.
For each test point $x$ let $\hat{x}$ denote the closest point in $\mathcal{P}$, $\hat{x}=\textrm{argmin}_{\mathcal{P}}\left(|x-z|\right)$. Adopting this notation, the result follows by the same argument as Theorem~\ref{thm:supp_interp_data}.
\end{proof}

\section{Random feature models}
Here we present random feature models in more detail. We begin by reviewing exact expressions for the loss. We then go onto derive its asymptotic properties. 
We again consider training a model $f(x)=\sum_{\mu=1}^{P}\theta_{\mu}f_{\mu}(x)$, where $f_{\mu}$ are drawn from some larger pool of features, $\{F_{M}\}$, $f_{\mu}(x)=\sum_{M=1}^{S}\mathcal{P}_{\mu M} F_{M}(x)$.

Note, if $\{F_{M}(x)\}$ form a complete set of functions over the data distribution, then any target function, $y(x)$, can be expressed as $y=\sum_{M=1}^{S}\omega_{M}F_{M}(x)$.
% , and we will adopt this notation below.
The extra constraint in a teacher-student model is specifying the distribution of the $\omega_{M}$. The variance-limited scaling goes through with or without the teacher-student assumption; however it is crucial for analyzing the resolution-limited behavior.

As in section 2.3 of the main text,
%~\ref{sec:kernel}
we consider models with weights initialized to zero and trained to convergence with mean squared error loss.
\es{eq:train_loss_sup}{
L_{\textrm{train}}&=\frac{1}{2D}\sum_{a=1}^{D}\left(f(x_{a})-y_{a}\right)^{2}\,.
}
The data and feature second moments play a central role in our analysis. We introduce the notation,
\es{eq:second_moment_def}{
&\mathcal{C}=\lexpp{x}F(x)F^{T}(x)\rexp\,, \ \ \ \bar{\mathcal{C}}=\frac{1}{D}\sum_{a=1}^{D}F(x_{a})F^{T}(x_{a})\,,
\ \ \ C=\mathcal{P}\mathcal{C}\mathcal{P}^{T}\,,
\ \ \ \bar{C}=\mathcal{P}\bar{\mathcal{C}}\mathcal{P}^{T}\,.\\
&\mathcal{K}(x,x')=\frac{1}{S}F^{T}(x)F(x')\,, \ \ \ \bar{\mathcal{K}}=\mathcal{K}\Big|_{\mathcal{D}_{\textrm{train}}}\,,
\ \ \ K(x,x')=\frac{1}{P}f^{T}(x)f(x')\,,
\ \ \ \bar{K}=K\Big|_{\mathcal{D}_{\textrm{train}}}\,.\\
}
Here the script notation indicates the full feature space while the block letters are restricted to the student features. The bar represents restriction to the training dataset. We will also indicate kernels with one index in the training set as $\vec{\cK}(x):=\cK(x,x_{a=1\ldots D})$ and $\vec{K}(x):=K(x,x_{a=1\ldots D})$. After this notation spree, the test loss can be written for \up{} models $P\leq D$ as
\es{eq:up_min}{
L(D,P)&=\frac{1}{2S}\lexpp{D}\trace\left(\mathcal{C}+ \bar{\cC}\mathcal{P}^{T}\bar{C}^{-1}C\bar{C}^{-1}\mathcal{P}\bar{\cC}
-2\bar{\cC}\cP^{T}\bar{C}^{-1}\cP\cC \right)\rexp\,,
}
and for \op{} models (at the unique minimum found by GD, SGD, or projected Newton's method),
\es{eq:op_min}{
L(D,P)=\frac{1}{2}\lexpp{x,D}\mathcal{K}(x,x)+\vec{K}(x)^{T}\bar{K}^{-1}\bar{\mathcal{K}}\bar{K}^{-1}\vec{K}(x)-2\vec{K}(x)^{T}\bar{K}^{-1}\vec{\mathcal{K}}(x)\rexp\,.
}
Here the expectation $\lexpp{D}\bullet\rexp$ is an expectation with respect to i.i.d. draws of a dataset of size $D$ from the input distribution, while $\lexpp{x}\bullet\rexp$ is an ordinary expectation over the input distribution. Note, expression \eqref{eq:up_min} is also valid for \op{} models and \eqref{eq:op_min} is valid for \up{} models if the inverses are replaces with the Moore-Penrose pseudo-inverse. Also note, the two expressions can be related by exchanging the projections onto finite features with the projection onto the training dataset and the sums of teacher features with the expectation over the data manifold. This realizes the duality between dataset and features discussed above.

\subsection{Asymptotic expressions}
We are interested in \eqref{eq:up_min} and \eqref{eq:op_min} in the limits of large $P$ and $D$.

\subsubsection*{Variance-limited scaling}
We begin with the \up{} case.
In the limit of lots of data, the sample estimate of the feature-feature second moment matrix, $\bar{\cC}$, approaches the true second moment matrix, $\cC$. Explicitly, if we define the difference $\delta\cC$ by $\bar{\cC}=\cC + \delta\cC$, we have
\es{eq:diff_scaling}{
\lexpp{D}\delta\cC\rexp&=0\\
\lexpp{D}\delta\cC_{M_{1}N_{1}}\delta\cC_{M_{2}N_{2}}\rexp&=\frac{1}{D}\left(\lexpp{x}F_{M_{1}}(x)F_{N_{1}}(x)F_{M_{2}}(x)F_{N_{2}}(x)\rexp-\cC_{M_{1}N_{1}}\cC_{M_{2}N_{2}}\right)\\
\lexpp{D}\delta\cC_{M_{1}N_{1}}\cdots\delta\cC_{M_{n}N_{n}}\rexp&=\mathcal{O}\left(D^{-2}\right) \ \ \forall n > 2\,.
}
The key takeaway from $\eqref{eq:diff_scaling}$ is that the dependence on $D$ is manifest.

Using these expressions in \eqref{eq:up_min} yields.

\es{eq:up_expansion}{
L(D,P)&=\frac{1}{2S}\trace\left(\cC-\cC\cP^{T}C^{-1}\cP\cC\right)\\
&+\frac{1}{2DS}\sum_{M_{1,2}N_{1,2}=1}^{P}T_{M_{1}N_{1}M_{2}N_{2}}\left[\delta_{M_{1}M_{2}}\left(\cP^{T}C^{-1}\cP\right)_{N_{1}N_{2}} + (C^{-1}\cP\cC^{2}\cP^{T}C^{-1})_{M_{1}M_{2}}C^{-1}_{N_{1}N_{2}}\right.\\
&\qquad\qquad\qquad\qquad\qquad\qquad\quad\left.-2\left(\cC\cP^{T}C^{-1}\cP\right)_{M_{1}M_{2}}\left(\cP^{T}C^{-1}\cP\right)_{N_{1}N_{2}} \right]+\mathcal{O}\left(D^{-2}\right)\,.
}
Here we have introduced the notation, $T_{M_{1}N_{1}M_{2}N_{2}}=\lexpp{x}F_{M_{1}}(x)F_{N_{1}}(x)F_{M_{2}}(x)F_{N_{2}}(x)\rexp$.

As above, defining 
\es{eq:up_min_simp_supp}{
L(P):=\lim_{D\rightarrow\infty}L(D,P)=\frac{1}{2S}\trace\left(\cC-\cC\cP^{T}C^{-1}\cP\cC\right)\,,
}
we see that though $L(D,P)- L(P)$ is a somewhat cumbersome quantity to compute, involving the average of a quartic tensor over the data distribution, its dependence on $D$ is simple.

For the \op{} case, we can similarly expand \eqref{eq:op_min} using $K=\cK+\delta\cK$. With fluctuations satisfying,
\es{eq:diff_scaling_op}{
\lexpp{P}\delta\cK\rexp&=0\\
\lexpp{P}\delta\cK_{a_{1}b_{1}}\delta\cK_{a_{2}b_{2}}\rexp&=\frac{1}{P}\left(\lexpp{P}f_{\mu}(x_{a_{1}})f_{\mu}(x_{b_{1}})f_{\mu}(x_{a_{2}})f_{\mu}(x_{b_{2}})\rexp-\cK_{a_{1}b_{1}}\cK_{a_{2}b_{2}}\right)\\
\lexpp{P}\delta\cK_{a_{1}a_{1}}\cdots\delta\cK_{a_{n}a_{n}}\rexp&=\mathcal{O}\left(P^{-2}\right) \ \ \forall n > 2\,.
}
This gives the expansion
\es{eq:op_expansion}{
L(D,P)&=\frac{1}{2}\lexpp{x,D}\mathcal{K}(x,x)-\vec{\cK}(x)^{T}\bar{\cK}^{-1}\vec{\cK}(x)\rexp + \mathcal{O}(P^{-1})\,,
}
and
\es{eq:op_min_simp_supp}{
L(D)=\frac{1}{2}\lexpp{x,D}\mathcal{K}(x,x)-\vec{\cK}(x)^{T}\bar{\cK}^{-1}\vec{\cK}(x)\rexp\,.}

\subsubsection*{Resolution-limited scaling} We now move onto studying the parameter scaling of $L(P)$ and dataset scaling of $L(D)$.
We explicitly analyze the dataset scaling of $L(D)$, with the parameter scaling following via the dataset parameter duality.

Much work has been devoted to evaluating the expression \eqref{eq:op_min_simp_supp} \citep{williams2000upper, NIPS2001_26f5bd4a, sollich2002learning}. One approach is to use the \emph{replica trick} -- a tool originating in the study of disordered systems which computes the expectation of a logarithm of a random variable via simpler moment contributions and analyticity assumptions \citep{parisi1980sequence}. The replica trick has a long history as a technique to study the generalization properties of kernel methods \citep{sollich1998learning, malzahn2001learning, malzahn2003learning, urry2012replica, cohen2019learning, gerace2020generalisation, bordelon2020spectrum}. We will most closely follow the work of \cite{canatar2020statistical}, who use the replica method to derive an expression for the test loss of linear feature models in terms of the eigenvalues of the kernel $\cC$ and $\bar{\omega}$, the coefficient vector of the target labels in terms of the model features.
\es{eq:canatar}{
&L(D)=\frac{\kappa^{2}}{1-\gamma}\sum_{i}\frac{\lambda_{i}\bar{\omega}_{i}^{2}}{\left(\kappa+D\lambda_{i}\right)^{2}}\,,\\ &\kappa=\sum_{i}\frac{\kappa\lambda_{i}}{\kappa+D\lambda_{i}}\,, \ \ \ \gamma=\sum_{i}\frac{D\lambda_{i}^{2}}{\left(\kappa+D\lambda_{i}\right)^{2}}\,.
}
This is the ridge-less, noise-free limit of equation (4) of \cite{canatar2020statistical}. Here we analyze the asymptotic behavior of these expressions for eigenvalues satisfying a power-law decay, $\lambda_i=i^{-(1+\alpha_{K})}$ and for targets coming from a teacher-student setup, $w\sim\cN(0,1/S)$.

To begin, we note that for teacher-student models in the limit of many features, the overlap coefficients $\bar{\omega}$ are equal to the teacher weights, up to a rotation $\bar{\omega}_{i}=O_{iM}w_{M}$. As we are choosing an isotropic Gaussian initialization, we are insensitive to this rotation and, in particular, $\lexpp{w}\bar{\omega}_{i}^{2}\rexp=1/S$. See Figure~\ref{fig:omega_bar} for empirical support of the average constancy of $\bar{\omega}_{i}$ for the teacher-student setting and contrast with realistic labels.

With this simplification, we now compute the asymptotic scaling of $\eqref{eq:canatar}$ by approximating the sums with integrals and expanding the resulting expressions in large $D$. We use the identities:
\es{eq:math_identities}{
\int_{1}^{\infty}dx\frac{x^{-n(1+\alpha)}}{\left(\kappa+D x^{-(1+\alpha)}\right)^{m}}&=\kappa^{-m}\frac{\Gamma\left(n-\frac{1}{1+\alpha}\right)}{(1+\alpha)\Gamma\left(n+\frac{\alpha}{1+\alpha}\right)}{}_2F_1\left(m,n-\frac{1}{1+\alpha},n+\frac{\alpha}{1+\alpha},\frac{-D}{\kappa}\right)\\
{}_2F_{1}\left(a,b,c,-y\right)&\propto y^{-a}+\mathcal{B}y^{-b}+\ldots\,.
}
Here ${}_2F_1$ is the hypergeometric function and the second line gives its asymptotic form at large $y$. $\mathcal{B}$ is a constant which does not effect the asymptotic scaling. 

Using these relations yields
\es{eq:replica_scaling_results}{
\kappa\propto D^{-\alpha_K},\quad \gamma\propto D^{0},\quad \text{and} \quad L(D)\propto D^{-\alpha_K}\,, 
}
as promised. Here we have dropped sub-leading terms at large $D$.
Scaling behavior for parameter scaling $L(P)$ follows via the data-parameter duality.

\begin{figure}[t]
     \centering
     \begin{subfigure}
         \centering
         \includegraphics[width=\textwidth]{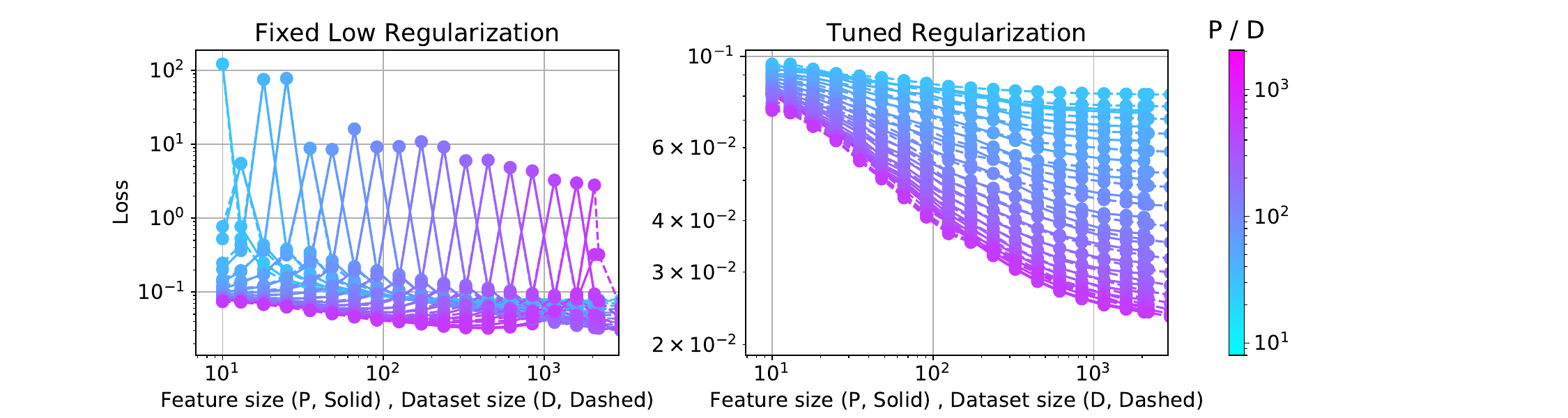}
     \end{subfigure}
        \caption{\textbf{Duality between dataset size vs feature number in pretrained features.} Using pretrained embedding features of EfficientNet-B5~\citep{tan2019efficientnet} for different levels of regularization, we see that loss as function of dataset size or loss as a  function of the feature dimension track each other both for small regularization (\textbf{left}) and for tuned regularization (\textbf{right}). Note that regularization strength with trained-feature kernels can be mapped to inverse training time~\citep{ali2019continuous, lee2020finite}. Thus (\textbf{left}) corresponds to long training time and exhibits double descent behavior, while (\textbf{right}) corresponds to optimal early stopping.} 
        \label{fig:Embedding_duality}
\end{figure}

\begin{figure}[t]
     \centering
     \begin{subfigure}
         \centering
         \includegraphics[width=0.47\textwidth]{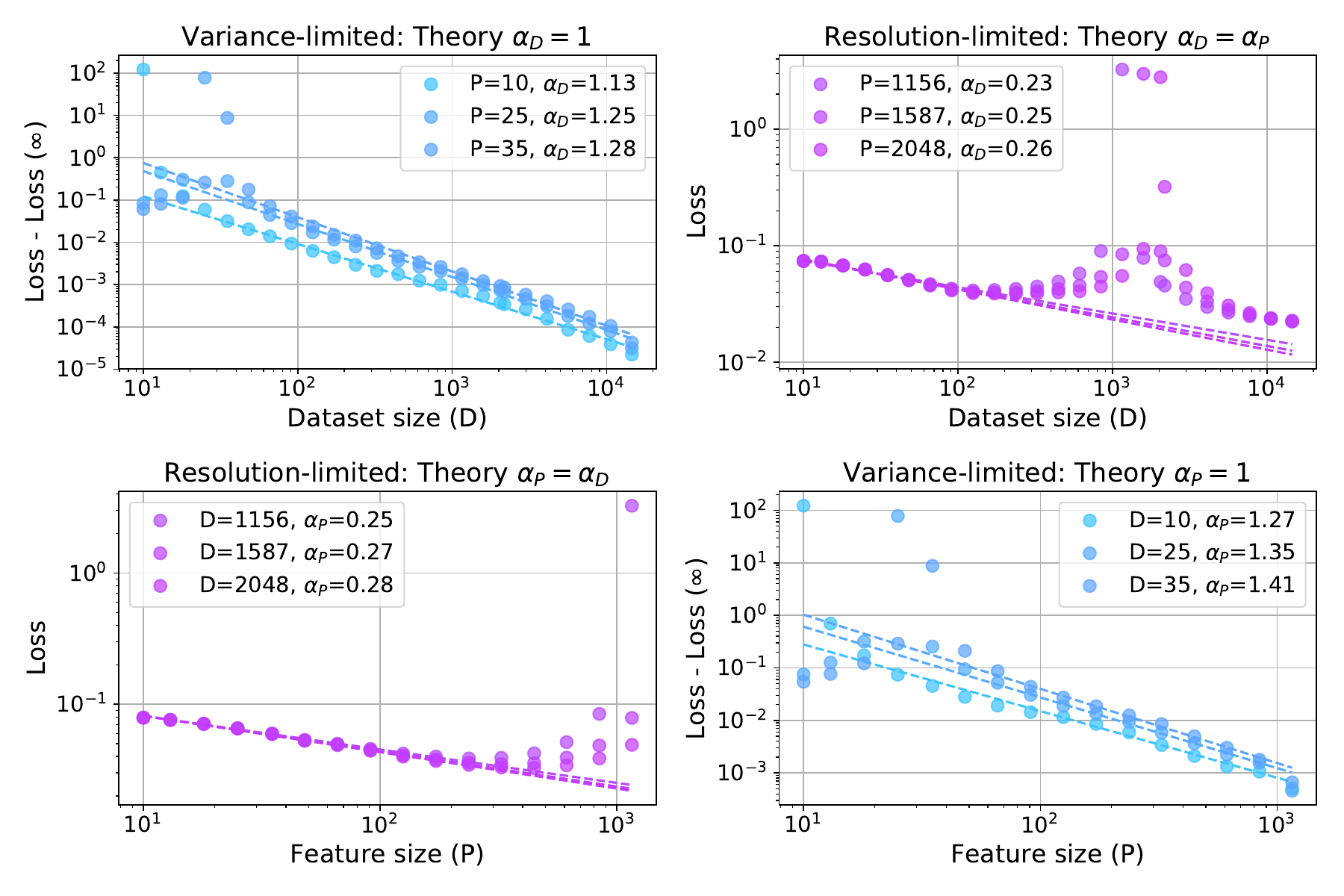}
         %\caption{}
     \end{subfigure}
     \begin{subfigure}
     \centering
     \includegraphics[width=0.47\textwidth]{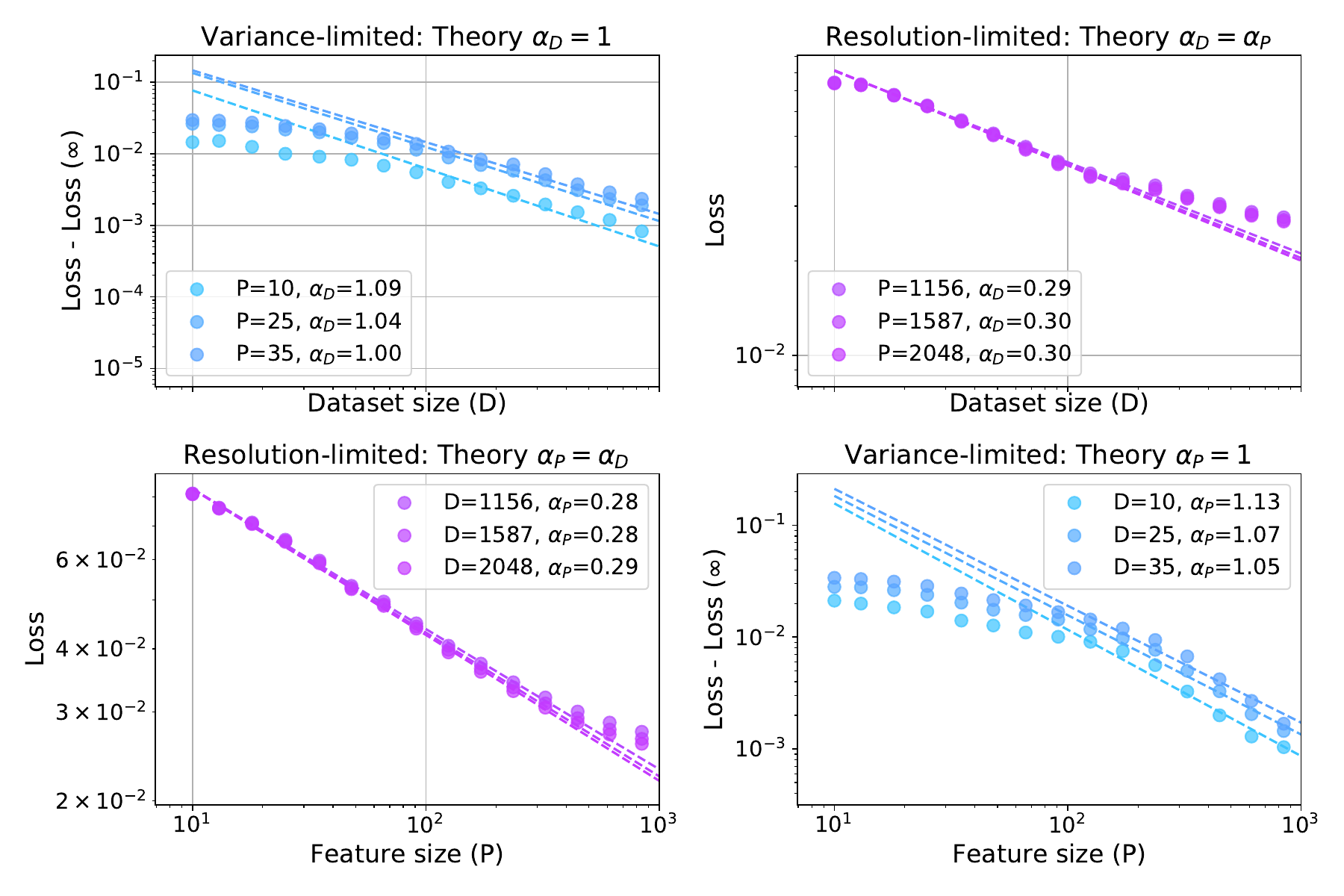}
         %\caption{}
     \end{subfigure}
     \hfill
        \caption{\textbf{Four scaling regimes exhibited by pretrained embedding features.} Using pretrained embedding features of EfficientNet-B5~\citep{tan2019efficientnet} for fixed low regularization (\textbf{left}) and tuned regularization (\textbf{right}), we can identify four regimes of scaling using real CIFAR-10 labels.}
        \label{fig:enet_duality}
\end{figure}

\subsection*{Duality beyond asymptotics}

Expressions \eqref{eq:up_min} and \eqref{eq:op_min} are related by changing projections onto finite feature set and finite dataset even without taking any asymptotic limits. We thus expect the dependence of test loss on parameter count and dataset size to be related quite generally in linear feature models. See Section~\ref{sec:learned features} for further details.

\section{Learned features}
\label{sec:learned features}
In this section, we consider linear models with features coming from pretrained neural networks. Such features are useful for transfer learning applications (e.g.~\cite{kornblith2019better, kolesnikov2019big}).
In Figures~\ref{fig:Embedding_duality} and \ref{fig:enet_duality}, we take pretrained embedding features from an EfficientNet-B5 model~\citep{tan2019efficientnet} using TF hub\footnote{https://www.tensorflow.org/hub}. The EfficientNet model is pretrained using the ImageNet dataset with input image size of $(456, 456)$. To extract features for the $(32, 32)$ CIFAR-10 images, we use \emph{bilinear} resizing.
We then train a linear classifier on top of the penultimate pretrained features. To explore the effect of feature size $P$ and dataset size $D$, we randomly subset the feature dimension and training dataset size and average over $5$ random seeds. Prediction on test points are obtained as a kernel ridge regression problem with linear kernel. We note that the regularization ridge parameter can be mapped to an inverse early-stopping time~\citep{ali2019continuous, lee2020finite} of a corresponding ridgeless model trained via gradient descent. Inference with low regularization parameter denotes training for long time while tuned regularization parameter is equivalent to optimal early stopping. 

In Figure~\ref{fig:enet_duality} we see evidence of all four scaling regimes for low regularization (left four) and optimal regularization (right four). We speculate that the deviation from the predicted variance-limited exponent $\alpha_P = \alpha_D = 1$ for the case of fixed low regularization (late time) is possibly due to the double descent resonance at $D = P$, which interferes with the power law fit.

In Figure~\ref{fig:Embedding_duality}, we observe the duality between dataset size $D$ (solid) and feature size $P$ (dashed) -- the loss as a function of the number of features is identical to the loss as a function of dataset size for both the optimal loss (tuned regularization) or late-time loss (low regularization). 

\begin{figure}[ht]
     \centering
     \begin{subfigure}
         \centering

        \includegraphics[width=0.95\textwidth]{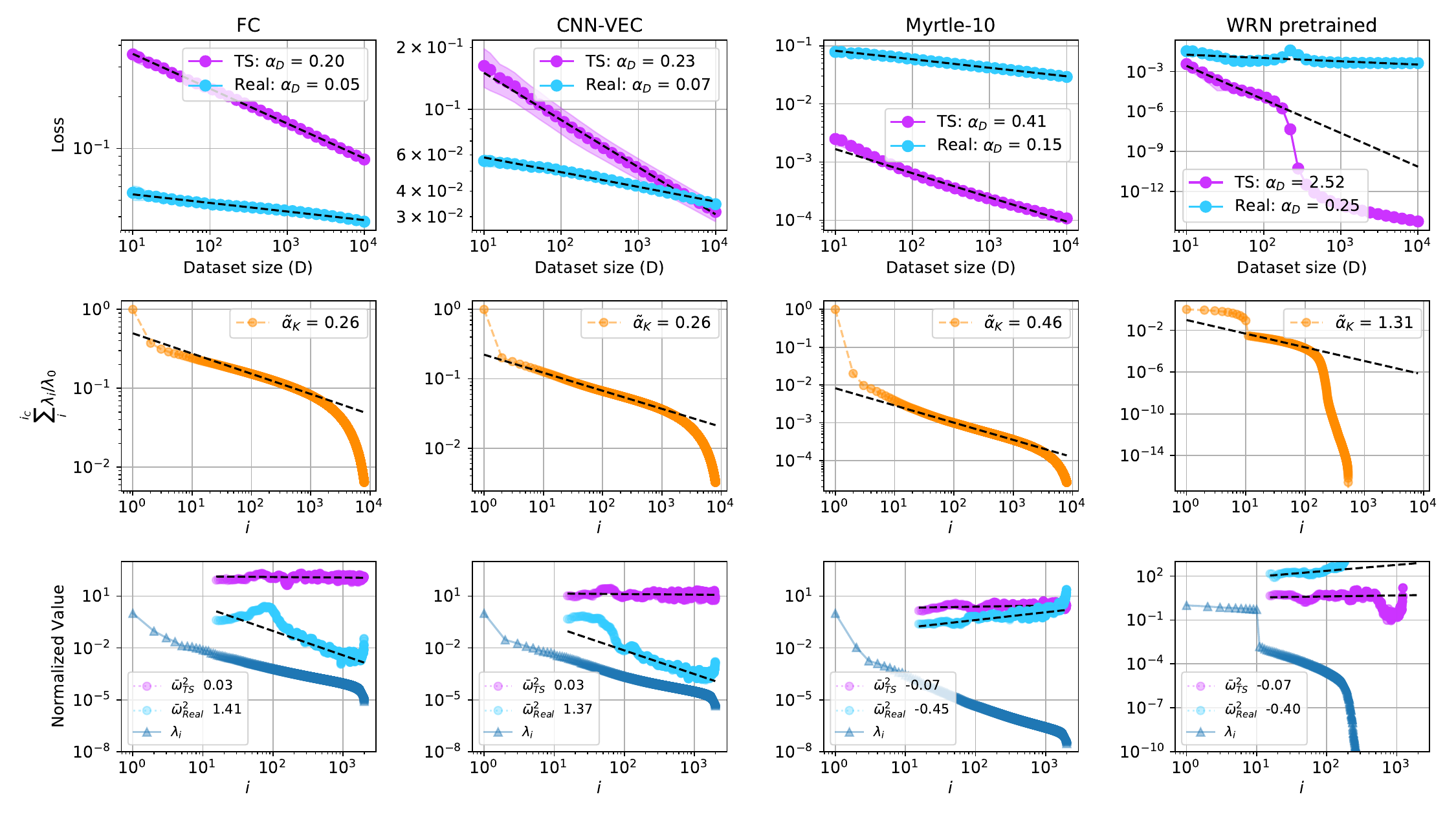}

    \end{subfigure}
        \caption{\textbf{Loss on the teacher targets scale better than real targets for both untrained and trained features.} The first three columns are infinite width kernels while the last column is a kernel built out of features from the penultimate layer of pretrained WRN 28-10 models on CIFAR-10. The first row is the loss as a function of dataset size $D$ for teacher-student targets vs real targets. The observed dataset scaling exponent is denoted in the legend. The second row is the normalized partial sum of kernel eigenvalues. The partial sum's scaling exponent is measured to capture the effect of the finite dataset size when empirical $\alpha_K$ is close to zero. The third row shows $\bar \omega_i$ for teacher-student  and real target compared against the kernel eigenvalue decay. We see the teacher-student $\bar{\omega}_{i}$ are approximately constant.} 
        \label{fig:omega_bar}
\end{figure}

In Figure~\ref{fig:omega_bar}, we also compare properties of random features (using the infinite-width limit) and learned features from trained Wide Resnet (WRN) 28-10 models.
We note that teacher-student models, where the feature class matches the target function and ordinary, fully trained models on real data (Figure 1a)
%~\ref{fig:four_regimes}a),
have significantly larger exponents than models with fixed features and realistic targets. 

The measured $\bar \omega_i$ -- the coefficient of the task labels under the $i$-th feature~\eqref{eq:canatar} -- are approximately constant as function of index $i$ for all teacher-student settings. However, for real targets, $\bar \omega_i$ are only constant for the well-performing Myrtle-10 and WRN trained features (last two columns).

\end{document}